\let\oldyear\year
\documentclass{ieeeaccess}

\let\year\oldyear
\usepackage{amsmath,amssymb,amsfonts}
\usepackage{graphicx}
\usepackage{textcomp}
\usepackage{float}
\usepackage[backend=biber,
sorting=none,
doi=false,
isbn=false,
url=true,
maxcitenames=5,
mincitenames=1,
style=ieee,]{biblatex}
\usepackage{multirow}
\usepackage{multicol}
\usepackage{booktabs}
\usepackage{colortbl}
\usepackage{bm}
\usepackage{graphicx}
\usepackage{algorithm}
\usepackage{algorithmicx} 
\usepackage{algpseudocode} 
\usepackage{etoolbox}
\usepackage{amsmath}
\usepackage{setspace}
\DeclareMathOperator*{\argmax}{argmax} 

\usepackage{tikz}
\NewSpotColorSpace{PANTONE}
\AddSpotColor{PANTONE} {PANTONE3015C} {PANTONE\SpotSpace 3015\SpotSpace C} {1 0.3 0 0.2}
\SetPageColorSpace{PANTONE}%
\usetikzlibrary{tikzmark}
\usepackage{pifont}
\newcommand{\xmark}{\ding{55}}%
\usetikzlibrary{calc}
\errorcontextlines\maxdimen

\newcommand{\ALGtikzmarkcolor}{black}
\newcommand{\ALGtikzmarkextraindent}{4pt}
\newcommand{\ALGtikzmarkverticaloffsetstart}{-.5ex}
\newcommand{\ALGtikzmarkverticaloffsetend}{-.5ex}
\makeatletter
\newcounter{ALG@tikzmark@tempcnta}

\newcommand\ALG@tikzmark@start{%
    \global\let\ALG@tikzmark@last\ALG@tikzmark@starttext%
    \expandafter\edef\csname ALG@tikzmark@\theALG@nested\endcsname{\theALG@tikzmark@tempcnta}%
    \tikzmark{ALG@tikzmark@start@\csname ALG@tikzmark@\theALG@nested\endcsname}%
    \addtocounter{ALG@tikzmark@tempcnta}{1}%
}

\def\ALG@tikzmark@starttext{start}
\newcommand\ALG@tikzmark@end{%
    \ifx\ALG@tikzmark@last\ALG@tikzmark@starttext
    \else
        \tikzmark{ALG@tikzmark@end@\csname ALG@tikzmark@\theALG@nested\endcsname}%
        \tikz[overlay,remember picture] \draw[\ALGtikzmarkcolor] let \p{S}=($(pic cs:ALG@tikzmark@start@\csname ALG@tikzmark@\theALG@nested\endcsname)+(\ALGtikzmarkextraindent,\ALGtikzmarkverticaloffsetstart)$), \p{E}=($(pic cs:ALG@tikzmark@end@\csname ALG@tikzmark@\theALG@nested\endcsname)+(\ALGtikzmarkextraindent,\ALGtikzmarkverticaloffsetend)$) in (\x{S},\y{S})--(\x{S},\y{E});%
    \fi
    \gdef\ALG@tikzmark@last{end}%
}

\apptocmd{\ALG@beginblock}{\ALG@tikzmark@start}{}{\errmessage{failed to patch}}
\pretocmd{\ALG@endblock}{\ALG@tikzmark@end}{}{\errmessage{failed to patch}}
\makeatother
\usepackage{amsmath}
\usepackage{amssymb}
\newcommand{\ALGORITHMHEADING}[1]{\item[\textcolor{red}{#1}]}
\makeatletter
\AtBeginDocument{\DeclareMathVersion{bold}
\SetSymbolFont{operators}{bold}{T1}{times}{b}{n}
\SetSymbolFont{NewLetters}{bold}{T1}{times}{b}{it}
\SetMathAlphabet{\mathrm}{bold}{T1}{times}{b}{n}
\SetMathAlphabet{\mathit}{bold}{T1}{times}{b}{it}
\SetMathAlphabet{\mathbf}{bold}{T1}{times}{b}{n}
\SetMathAlphabet{\mathtt}{bold}{OT1}{pcr}{b}{n}
\SetSymbolFont{symbols}{bold}{OMS}{cmsy}{b}{n}
\renewcommand\boldmath{\@nomath\boldmath\mathversion{bold}}}
\makeatother

\def\BibTeX{{\rm B\kern-.05em{\sc i\kern-.025em b}\kern-.08em
    T\kern-.1667em\lower.7ex\hbox{E}\kern-.125emX}}

\begin{document}
\newcolumntype{P}[1]{>{\centering\arraybackslash}p{#1}}
\history{Date of publication xxxx 00, 0000, date of current version xxxx 00, 0000.}
\doi{10.1109/ACCESS.2024.0429000}

\title{Lifelong Whole Slide Image Analysis: Online Vision-Language Adaptation and Past-to-Present Gradient Distillation}
\author{\uppercase{Doanh C. Bui}$^{1}$,
\uppercase{Hoai Luan Pham}$^{1}$, \IEEEmembership{Member, IEEE}, \uppercase{Vu Trung Duong Le}$^{1}$, \IEEEmembership{Member, IEEE},
\uppercase{Tuan Hai Vu}$^{1}$, \IEEEmembership{Member, IEEE},
\uppercase{Van Duy Tran}$^{1}$,
\uppercase{Khang Nguyen}$^{2}$,
and \uppercase{Yasuhiko Nakashima}$^{1}$, \IEEEmembership{Senior Member, IEEE}}

\address[1]{Nara Institute of Science and Technology, Ikoma 630-0192, Japan}
\address[2]{University of Information Technology, Ho Chi Minh city Vietnam National University}

\markboth
{Bui \headeretal: Lifelong Whole Slide Image Analysis: Online Vision-Language Adaptation and Past-to-Present Gradient Distillation}
{Bui \headeretal: Lifelong Whole Slide Image Analysis: Online Vision-Language Adaptation and Past-to-Present Gradient Distillation}

\corresp{Corresponding author: Doanh C. Bui (e-mail: bui.cao\_doanh.bd2@naist.ac.jp).}

\begin{abstract}
Whole Slide Images (WSIs) play a crucial role in accurate cancer diagnosis and prognosis, as they provide tissue details at the cellular level. However, the rapid growth of computational tasks involving WSIs poses significant challenges. Given that WSIs are gigapixels in size, they present difficulties in terms of storage, processing, and model training. Therefore, it is essential to develop lifelong learning approaches for WSI analysis. In scenarios where slides are distributed across multiple institutes, we aim to leverage them to develop a unified online model as a computational tool for cancer diagnosis in clinical and hospital settings.
In this study, we introduce ADaFGrad, a method designed to enhance lifelong learning for whole-slide image (WSI) analysis. First, we leverage pathology vision–language foundation models to develop a framework that enables interaction between a slide’s regional tissue features and a predefined text-based prototype buffer. Additionally, we propose a gradient‐distillation mechanism that mimics the gradient of a logit with respect to the classification-head parameters across past and current iterations in a continual‐learning setting.
We construct a sequence of six TCGA datasets for training and evaluation. Experimental results show that ADaFGrad outperforms both state-of-the-art WSI-specific and conventional continual-learning methods after only a few training epochs, exceeding them by up to $+5.068\%$ in the class-incremental learning scenario while exhibiting the least forgetting (i.e., retaining the most knowledge from previous tasks). Moreover, ADaFGrad surpasses its baseline by as much as $+40.084\%$ in accuracy, further demonstrating the effectiveness of the proposed modules.
\end{abstract}


\titlepgskip=-21pt

\maketitle

\section{Introduction}
\label{sec:introduction}

In recent years, with the development of deep learning, digital computational tools have been created to analyze Whole Slide Images (WSIs) to support cancer diagnosis and prognosis \cite{wu2023artificial}. WSIs are large and show tissues at the cellular level, which helps pathologists identify abnormal regions and make more precise diagnoses than with other traditional methods \cite{kiran2023digital}. However, as WSIs are gigasized, they require significant storage, processing time, and transfer time between devices. With the rapid increase in WSI volume every day, WSI-related tasks for diagnostic support also grow. Consequently, developing computational tools, such as training new deep neural networks for new tasks, is time-consuming and resource-intensive. Furthermore, slides may be distributed among institutes and differ in terms of scanning processes \cite{jahanifar2023domain} (e.g., stains or different scanner settings), requiring the model to have the ability to update continually. Meanwhile, in current developments, we still need to retrain the model and utilize it in a distributed manner. 

\begin{figure}[http]
\centerline{\includegraphics[width=0.49\textwidth]{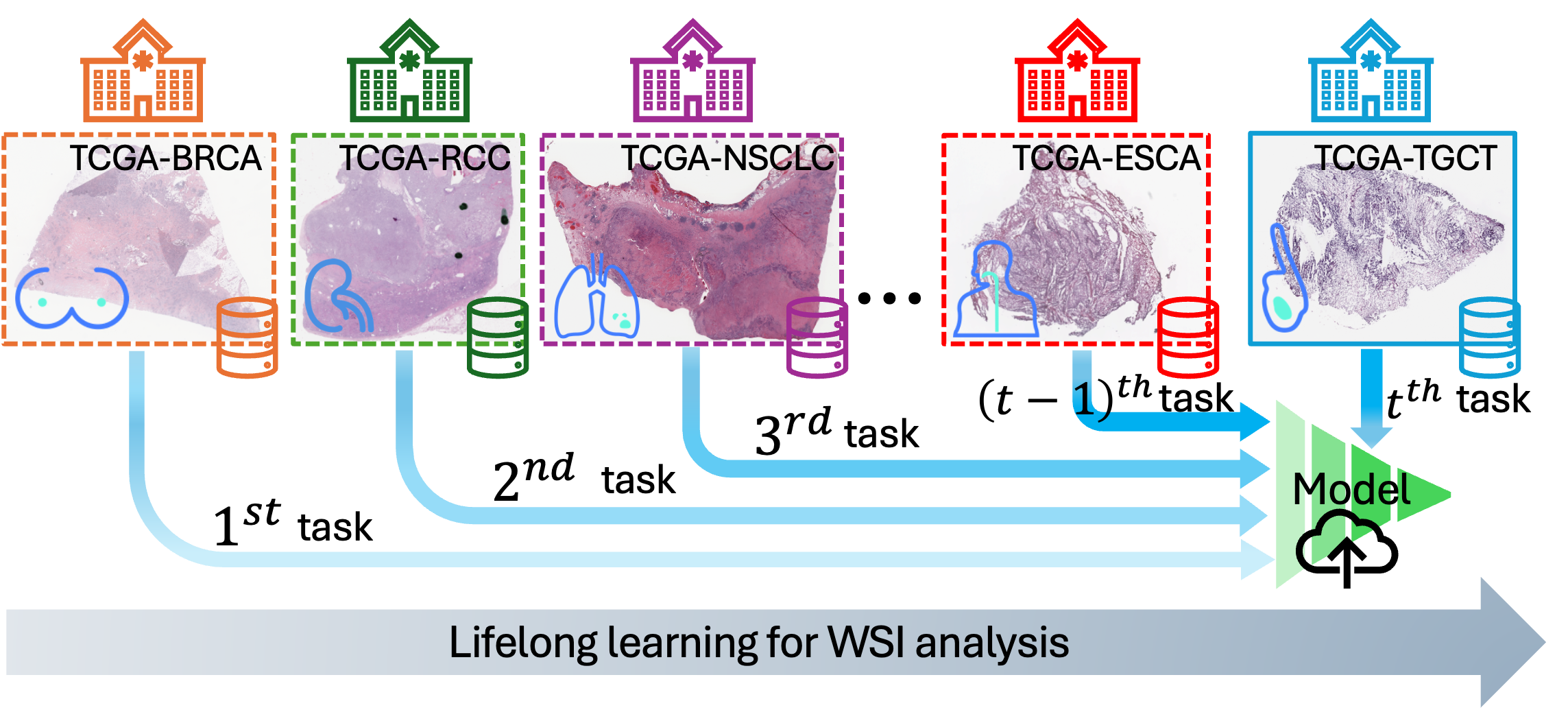}}
\caption{Lifelong learning for WSI analysis.}
\label{fig:thumbnail} 
\end{figure}

Given this problem, it is necessary to develop a framework that helps a unified model expand into new tasks without retraining on past tasks or training a new model for each task, referred as lifelong learning for WSI analysis (illustrated in Fig. \ref{fig:thumbnail}). Naive fine-tuning of a model with a new task may lead to catastrophic forgetting \cite{li2019learn,huang2023conslide}, where the model's parameter space significantly drifts to accommodate the new task and almost forgets how to predict on old tasks. Early studies attempted to constrain the parameter space to stay near the minimal solution of past tasks to avoid forgetting, by designing a regularization term \cite{li2017learning, kirkpatrick2017overcoming}. Later approaches relaxed the continual learning approach by allowing the model, when trained on new tasks, to see a few samples in an online-update buffer, referred to as replay. These approaches are categorized as rehearsal-based methods, which focus on designing different types of replay. Leveraging the rehearsal-based approach, \citeauthor{huang2023conslide} \cite{huang2023conslide} proposed diversifying the buffer specifically for WSIs by storing tissue regions rather than entire slides, and randomly sampling regions to form a slide for replay. In the context of continual learning for WSIs, \citeauthor{zhu2024lifelong} \cite{zhu2024lifelong} introduced a distance consistency loss to help slide representations remain close in the embedding space to slides in the online buffer for continual learning retrieval tasks. Meanwhile, these WSI-specific continual learning studies simply propose a strategy for storing samples in the buffer or introduce a loss function, while neglecting important aspects, such as the gradient of the logit with respect to the network's parameters. Carefully controlling the gradient may help prevent the model from shifting too far from the previously learned tasks when adding new ones \cite{erace,agem}. 

With the development of pathology vision-language foundation models, such as CONCH \cite{lu2024visual} and TITAN \cite{ding2024multimodal}, which train a vision encoder (e.g., Vision Transformer) and a text encoder to align with each other under a self-supervised learning framework using large amounts of image-text pairs, leveraging these models could enhance performance. MI-Zero \cite{lu2023visual} demonstrated that vision-text similarity computation can effectively perform zero-shot classification for WSIs. \citeauthor{huang2024free} \cite{huang2024free} introduced CATE, a method to maximize the informative features of a bag of patch features by aligning them with positive concept anchors, defined as corresponding class templates, and minimizing superfluous information by applying Kullback-Leibler Divergence loss to patch features against a negative set of texts that are irrelevant to the slide. However, these studies also introduced computational costs for text embedding during both training and inference, which may not be applicable in situations where resources and processing time need to be optimized as much as possible.

In this study, we attempt to design an effective lifelong learning method for WSI analysis. First, we leverage a pathology vision-language foundation model to extract a set of prototype texts. The set of prototype texts is grouped into two categories: class prototypes and negative prototypes. Class prototypes include sentences describing cancer subtyping classes for a task, while negative prototypes include irrelevant information. During training, we introduce Online Vision-Language Adaptation (OVLA), a loss function that pulls the bag of patch features of a slide towards its corresponding class prototype embedding and pushes away others, including different class prototype embeddings and negative prototype embeddings. Because this process is performed only during training and not during inference, we use the term ``online.'' This ensures that no additional computational cost is introduced during testing. To further preserve knowledge memory as the number of tasks increases, we access the gradient of the produced logits with respect to the parameters of classification head and introduce Past-to-Present Gradient Distillation (PPGD) to mimic the gradient of the model in the current iteration with that from the past iteration, helping to avoid forgetting. Leveraging both OVLA and PPGD forms our framework: \textbf{Ada}pting Pathology \textbf{F}oundation Vision-Language Model and Past-to-Present \textbf{Grad}ient Distillation, abbreviated as \textbf{ADaFGrad}. For evaluation, we establish a sequence of six TCGA datasets, covering six cancer subtyping tasks for breast, kidney, lung, esophagus, testis, and cervix uteri. \textbf{The experimental results show that our ADaFGrad surpasses state-of-the-art methods for lifelong learning in WSI analysis, both quantitatively and qualitatively, while requiring significantly fewer training epochs than other methods.}

\section{Related Works}
\label{sec:relatedworks}
\noindent\textbf{Multiple Instance Learning Models.} Multiple Instance Learning (MIL) is a research area that aims to produce an outcome for a bag of instances rather than for individual instances \cite{chen2006miles,sudharshan2019multiple}. A bag is considered to be assigned a positive label if at least one instance is positive. This is well aligned with WSIs, where they are typically treated as bags of patches. If only one patch appears abnormal, the entire slide should be identified as abnormal. Recent advances in MIL have shown to enhance WSI analysis. Early methods like ABMIL \cite{ilse2018attention} used attention to identify key patches, while CLAM \cite{clam} refined this by focusing on both highly and minimally contributory patches. TransMIL \cite{transmil} introduced ViT-based modeling to capture long-range dependencies. To address limited data, DTFD-MIL \cite{dtfd} used a multi-stage design with sub-MIL for better patch embeddings.

\noindent\textbf{Continual Learning Methods.} 
There have been various continual learning approaches applied to image classification tasks. For instance, Elastic Weight Consolidation (EwC) \cite{kirkpatrick2017overcoming} leveraged parameters from the previous task’s model to regularize training of the current model. GDumb \cite{gdumb} prevented forgetting by retraining a new model using a balanced memory buffer. ER-ACE \cite{erace} reweighted losses during experience replay to balance new and previous tasks. A-GEM \cite{agem} prevented interference among tasks by constraining gradients from new tasks. DER++ \cite{derpp} enhanced replay by using distillation to align current predicted logits with past logits. ConSlide \cite{huang2023conslide} made the first attempt to develop continual learning for WSI analysis using a breakup-reorganize strategy to increase the diversity of slides stored in the buffer.

\begin{figure*}[h!]
\centerline{\includegraphics[width=0.9\textwidth]{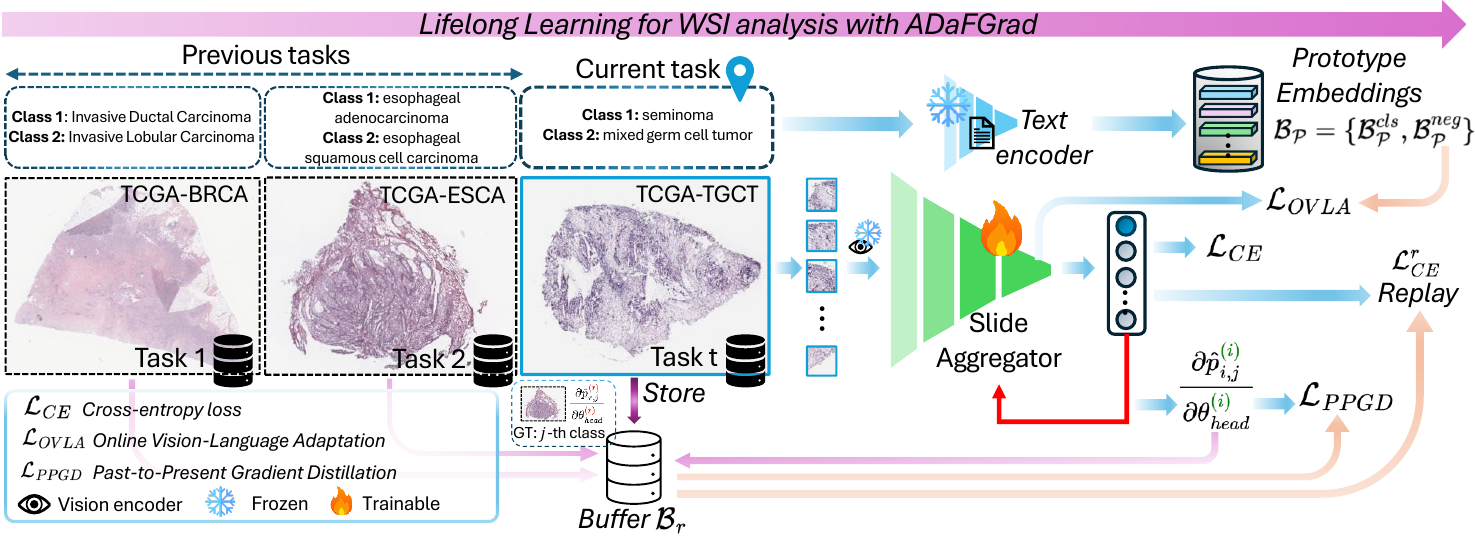}}
\caption{\textbf{Overview of ADaFGrad:} For a new task, class templates and irrelevant sentences serve as prototypes. These are encoded by a text encoder to extract prototype embeddings, which are stored in a prototype buffer. When a slide is processed, it is supervised by a cross-entropy loss $\mathcal{L}_{CE}$ and aligned with prototypes using the Online Vision-Language Adaptation loss $\mathcal{L}_{OVLA}$. Meanwhile, the gradient of the logit for the target label with respect to the classification head's parameters is stored and replayed for knowledge distillation through $\mathcal{L}_{PPGD}$.}
\label{fig:overview} 
\end{figure*}

\noindent\textbf{Pathology Foundation Models \& Adaptation.} 
Recent efforts have focused on developing foundation models for pathology image analysis, using CNN- or ViT-based architectures pretrained on large public or private datasets via self-supervised methods like DINO \cite{caron2021emerging} and MoCo v3 \cite{chen2021empirical}. Notable models include UNI \cite{chen2024towards} and ProvGigaPath \cite{xu2024whole}, though these are trained solely on image data. In contrast, multimodal foundation models, such as CONCH \cite{lu2024visual} and TITAN \cite{ding2024multimodal}, leverage vision-text pairs. To enhance the patch features and make them more discriminative for prediction, CATE \cite{huang2024free} used CONCH to improve patch features for better performance. Nonetheless, it introduced additional costs for text embedding extraction and integration during both training and inference.

\section{Method: ADaFGrad}
\label{sec:adafgrad}

\subsection{Problem Definition} We address the problem of lifelong learning for whole slide image analysis. Given a sequence of $N_t$ datasets corresponding to $N_t$ tasks, denoted as
$
\mathcal{D} = \left\{\left(D^{\text{train}}_i, D^{\text{test}}_i\right) \right\}_{i=1}^{N_t},
$ where each dataset includes $C_i$ cancer subtypes, we develop \textbf{ADaFGrad} to enable a unified model $\mathcal{F}$ to be sequentially trained and evaluated on each $D^{\text{train}}_t$. The key constraint is that after being trained on $D^{\text{train}}_i$, the model $\mathcal{F}$ should retain its performance and not forget previously learned datasets $\{D^{\text{test}}_i\}_{i<t}$. Following the evaluation settings of continual learning studies \cite{agem,derpp,huang2023conslide}, there are two scenarios: \textbf{class-incremental (CLASS-IL)} and \textbf{task-incremental (TASK-IL)}. CLASS-IL requires the model to correctly predict the true class label across all accumulated classes as the number of tasks grows, whereas TASK-IL considers only the logits for the current task’s classes. Hence, CLASS-IL is more challenging.

\subsection{Framework Overview}

An overview of ADaFGrad is illustrated in Fig. \ref{fig:overview}. \textbf{ADaFGrad} is built on a rehearsal-based continual learning framework, where the model can revisit a few samples stored in an online, fixed-size buffer $\mathcal{B}_r$. During training, samples from the buffer are randomly replayed. A WSI from dataset $D^{train}_t$ of the $t$-th task undergoes a tiling and patch feature extraction process. It then passes through the unified model $\mathcal{F}$ for cancer subtype prediction. The unified model $\mathcal{F}$ includes a slide aggregator $f_{\mathcal{A}}$ and a classification head $f_{head}$ to produce logits for prediction. We adopt \textbf{HIT}~\cite{huang2023conslide} as $f_{\mathcal{A}}$, a hierarchical transformer that leverages the pyramid structure of WSIs. ADaFGrad introduces two key loss functions: (1) \textit{Online Vision-Language Adaptation} ($\mathcal{L}_{OVLA}$) and (2) \textit{Past-to-Present Gradient Distillation} ($\mathcal{L}_{PPGD}$). $\mathcal{L}_{OVLA}$ aligns visual patch features with pre-defined text-based prototype embeddings to produce more discriminative representations. $\mathcal{L}_{PPGD}$ mimics the gradient of the logit (w.r.t. the classification head) from the current model and its previous version to mitigate catastrophic forgetting.

\subsection{WSI Tiling \& Feature Extraction}

Given a WSI $s$, we tile it into $N_r$ regions using the segmentation and patching strategy of CLAM~\cite{clam}. We then use TITAN’s vision encoder $f^{\text{TITAN}}_{\text{vision}}$ \cite{ding2024multimodal} to extract features for each region. This yields a sequence of region features $\mathcal{R} = \{r_i\}_{i=1}^{N_r}$, where $r_i \in \mathbb{R}^{C_{\mathrm{vis}}}$ is a $C_{\mathrm{vis}}$-dimensional feature vector. For each region, we further tile it into $k \times k$ patches, which also undergo $f^{\text{TITAN}}_{\text{vision}}$ to obtain a set of patch features $\mathbf{x}_i = \{x_j\}^{k \times k}_{j=1}$ for each region. For $N_r$ regions, we obtain $\mathcal{X} = \{\mathbf{x}_i\}^{N_r}_{i=1}$. The sequence of regions $\mathcal{R}$ and corresponding patches $\mathcal{X}$ then undergo a slide aggregation function $f_{\mathcal{A}}$ to obtain a single slide embedding. For convenience, we denote the $i$-th slide in a dataset as $s_i = \{\mathcal{R}_i, \mathcal{X}_i\}$ for an easier description of the operation flow in the following sections.

\subsection{Adaptation Of Foundation Vision-Language Model} This section describes in detail $\mathcal{L}_{OVLA}$. During training sequence of tasks continually, there is an online text-based prototype buffer $\mathcal{B}_{\mathcal{P}}$. Whenever $t$-th new task comes in, we define a set of class prototypes $\mathcal{P}^{cls}_t$ and negative prototypes $\mathcal{P}^{neg}_t$. 

\subsubsection{Creation Process of Text-based Prototypes}

\noindent For \textbf{Class Prototypes}, $\mathcal{P}^{cls}_t$ includes sentences describe class information for $t$-th cancer subtype task. Given $t$-th task including $C_t$ cancer subtyping classes, we define $T=22$ base templates (e.g., \texttt{``a histopathological image showing [CLASS].}'') and generate $\approx4$ phrasing variants per class. For each $j$-th class of $t$-th task, the set of class sentences is denoted as $\{cls^{j,t}_k\}^{4T}_{k=1}$.

\noindent For \textbf{Negative Prototypes}, We define a set of $l$ negative prototypes $\mathcal{P}^{neg}_t$, each containing a short description of irrelevant content (e.g., \textit{adipocytes}, \textit{connective tissue}, or \textit{normal tissue cells}). These descriptions do not contribute significantly to cancer prediction. By applying $T = 22$ class templates to each of the $l$ negative prototypes, we obtain a total of $22l$ negative sentences. For each $k$-th among $l$ negative prototypes of the $t$-th task, the set of negative sentences is denoted as $\{ neg^{t}_k \}_{k=1}^{T}$.

\begin{figure}[http]
\centerline{\includegraphics[width=0.49\textwidth]{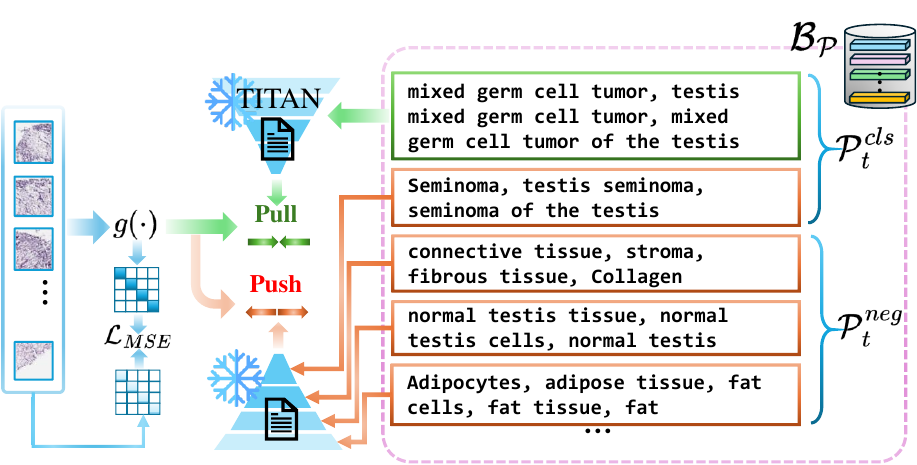}}
\caption{Illustration of OVLA.}
\label{fig:dataset} 
\end{figure}

\noindent For both sentences in $\mathcal{P}^{cls}_t$ and $\mathcal{P}^{neg}_t$, they undergo a text encoder of the TITAN model, $f^{\text{TITAN}}_{\text{text}}$, to obtain embeddings. As mentioned, for each class in $\mathcal{P}^{cls}_t$, there are $\approx 4T$ class sentences. All will undergo $f^{\text{TITAN}}_{\text{text}}$, obtain embeddings, and apply average pooling to form a single class prototype embedding $c_j$ for the $j$-th cancer subtyping class. Hence, $\mathcal{P}^{cls}_t=\{p^{cls}_j\}_{j=1}^{C_t}$. Similarly, we obtain a set of negative prototype embeddings $\mathcal{P}^{neg}_t=\{p^{neg}_j\}_{j=1}^{l}$. Both $p^{cls}_j,p^{neg}_j \in \mathbb{R}^{C^{\text{text}}}$. As the number of tasks increases, we add $\mathcal{P}^{cls}_t$ and $\mathcal{P}^{neg}_t$ to the text-based prototype buffer $\mathcal{B}_{\mathcal{P}}$:

\begin{equation}
\mathcal{B}^{cls}_{\mathcal{P}} \leftarrow \mathcal{B}^{cls}_{\mathcal{P}} \cup \mathcal{P}^{cls}_t, \quad \mathcal{B}^{neg}_{\mathcal{P}} \leftarrow \mathcal{B}^{neg}_{\mathcal{P}} \cup \mathcal{P}^{neg}_t.
\end{equation}

\subsubsection{Training Procedure of Online Vision-Language Adaptation}

\noindent During training, from the last backbone layer before obtaining the single slide embedding, we obtain a set of region features $\mathbf{z}=\{z_i\}^{N_r}_{i=1}$, where $z_i \in \mathbb{R}^{d_{\text{model}}}$ is a feature vector for a tissue region. \includegraphics[height=2ex]{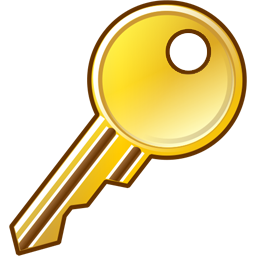}  \textbf{We aim to push the region features $\mathbf{z}$ of a WSI, whose ground-truth label corresponds to the $j$-th cancer subtyping class, towards its corresponding prototype embedding $p^{cls}_j$ and pull it away from all other prototype embeddings in a contrastive learning manner.} To achieve this, we adopt the InfoNCE loss function \cite{oord2018representation}, denoted as $\mathcal{L}_{InfoNCE}$ for each region in $\mathbf{z}$:

\begin{equation}
-\mathbb{E}_{g(z_i)}\left[
\log
\frac{
f\big(g(z_i), p^{cls}_j\big)
}{
\sum_{k \neq j}^{|\mathcal{B}^{cls}_{\mathcal{P}}|} f\big(g(z_i), p^{cls}_k\big)
+ \sum_{k=1}^{|\mathcal{B}^{neg}_{\mathcal{P}}|} f\big(g(z_i), p^{neg}_k\big)
}
\right]
\label{eq:infonce}
\end{equation}

\noindent where $f(\cdot)$ denotes the scoring function, and the exponential function $\exp(\cdot)$ is used. Specifically, $f(z_i, p) = \exp(\tau \cdot z_i^\intercal p)$. To avoid feature collapsing, diversify the feature values, and map the model's dimension to the prototype embedding, an alignment network $g(\cdot): \mathbb{R}^{d_{\text{model}}} \rightarrow \mathbb{R}^{C_{\text{text}}}$ is used. However, Eq. \ref{eq:infonce} still indirectly enhances $z_i$ through the alignment network $g(\cdot)$. To further ensure $z_i$ fully benefits from Eq. \ref{eq:infonce}, we design a mechanism to minimize the difference between $z_i$ and $g(z_i)$. Since they have different dimensional sizes, a self-similarity supervision term is applied:

\begin{equation}
\mathcal{L}_{sim} = \left\| g(\mathbf{z})^{\intercal} g(\mathbf{z}) - \mathbf{z}^{\intercal} \mathbf{z} \right\|^2_2.
\end{equation}

\noindent Finally, Online Vision-Language Adaptation is formulated as:

\begin{equation}
\mathcal{L}_{OVLA} = \alpha\mathcal{L}_{InfoNCE} + \beta\mathcal{L}_{sim},
\label{eq:ovla}
\end{equation}

\noindent where $\alpha$ and $\beta$ are hyperparameters that control the contribution of each loss term.

\subsection{Past-to-Present Gradient Distillation} This section describes in detail $\mathcal{L}_{PPGD}$. We attempt to access and control the gradient of the produced logits with respect to the classification head of $f_{head}(\cdot, \theta_{head})$, where $\theta_{head} \in \mathbb{R}^{d_{\text{model}} \times \mathcal{C}}$ and $\mathcal{C}=\sum_{i=1}^{N_t} |C_i|$, is the set of learnable parameters for the classification head. Given a sequence of regions $\mathcal{R}_i$ and patches $\mathcal{X}_i$ for a WSI $s_i$, we obtain the produced logits from the model at $i$-th iteration:

\begin{equation}
\hat{p}^{(i)}_i = f^{(i)}_{head}\left(f_{\mathcal{A}}^{(i)}(\mathcal{R}_i, \mathcal{X}_i), \theta^{(i)}_{head}\right),
\end{equation}

\noindent where $\hat{p}^{(i)}_i \in \mathbb{R}^{\mathcal{C}}$ is the logit prediction for the $i$-th WSI at the $i$-th iteration model. The cross-entropy loss function is then used for conventional classification on sample $s_i$ with the ground-truth cancer subtyping label $y_i$ at one-hot encoding form:

\begin{equation} \mathcal{L}_{CE}(\hat{p}^{(i)}_i, y_i) = -\sum^{\mathcal{C}}_{j=1} y_{i,j}\log(\hat{p}^{(i)}_{i,j}). \end{equation}

\noindent Assuming a WSI $s_i$ belongs to the $j$-th cancer subtyping class, we compute the logit for the $j$-th class label with respect to the parameters of the classification head at the current iteration $i$-th, $\theta^{(i)}_{head}$, and store it in the buffer $\mathcal{B}_r$ along with $s_i$:

\begin{equation}
\mathcal{B}_r \leftarrow \mathcal{B}_r \cup^* \left\{\left((\mathcal{R}_i,\mathcal{X}_i), \frac{\partial \hat{p}^{(i)}_{i,j}}{\partial \theta^{(i)}_{head}}\right)\right\}, \quad \text{where } \frac{\partial \hat{p}^{(i)}_{i,j}}{\partial \theta^{(i)}_{head}} \in\mathbb{R}^{d_{\text{model}}}.
\end{equation}

\noindent where $\cup^*$ is a conditional addition operation. It will add the new sample to the buffer $\mathcal{B}_r$ if the buffer size does not exceed the pre-defined fixed size $|\mathcal{B}_r|$; otherwise, the new sample will randomly replace old samples in the buffer. Then, whenever the model performs \textit{replay}, i.e., when the $i$-th iteration model is trained on a randomly sampled slide $s_r$ from the buffer $\mathcal{B}_r$, we additionally perform past-to-present gradient distillation. \includegraphics[height=2ex]{figures/key.png}  \textbf{Specifically, we minimize the similarity between the gradient of the current $i$-th iteration model and the past $r$-th gradient on the same WSI $s_r$.} First, we perform \textit{replay}:

\begin{equation}
\hat{p}^{(i)}_r = f^{(i)}_{head}\left(f_{\mathcal{A}}^{(i)}(\mathcal{R}_r, \mathcal{X}_r), \theta^{(i)}_{head}\right), \text{ where } (\mathcal{R}_r,\mathcal{X}_r) \in \mathcal{B}_r
\end{equation}

\begin{equation}
\mathcal{L}^r_{\text{CE}}(\hat{p}^{(i)}_r, y_r) = -\sum^{\mathcal{C}}_{j=1} y_{r,j} \log(\hat{p}^{(i)}_{r,j}),
\end{equation}

\noindent where $\hat{p}^{(i)}_r$ is the logit prediction for the $r$-th WSI sampled from the buffer by the $i$-th iteration model. After that, we compute the gradient of the classification head's parameters at the $i$-th iteration with respect to $\hat{p}^{(i)}_r$, i.e., ${\partial \hat{p}^{\textcolor{green}{(i)}}_{\textcolor{red}{r},j}}/{\partial \theta^{\textcolor{green}{(i)}}_{head}}$, and minimize it with the old $r$-th gradient stored in the buffer via the following loss function:

\begin{equation}
\mathcal{L}_{PPGD} = 1 - \left( \frac{\partial \hat{p}^{\textcolor{red}{(r)}}_{\textcolor{red}{r},j}}{\partial \theta^{\textcolor{red}{(r)}}_{head}} \right)^{\intercal} \left( \frac{\partial \hat{p}^{\textcolor{green}{(i)}}_{\textcolor{red}{r},j}}{\partial \theta^{\textcolor{green}{(i)}}_{head}} \right).
\label{eq:ppgd}
\end{equation}

\subsection{Final Training Objective}

Finally, the training objective for each WSI sample is defined as:

\begin{equation} \min_{\theta_{\mathcal{F}}} \mathcal{L}_{CE} + \mathcal{L}_{OVLA} + \gamma\mathcal{L}^r_{CE} + \tau\mathcal{L}_{PPGD}. \end{equation}

\noindent where $\theta_{\mathcal{F}} = \{{\theta_{\mathcal{A}}, \theta_{head}}\}$ are the parameters for the unified model $\mathcal{F}$, and $\gamma$ and $\tau$ are hyperparameters that control the contribution of each loss term. For summarization of ADaFSlide, we provide Algorithm \ref{alg_1}.

\begin{algorithm}
\setstretch{0.92}
\footnotesize
\caption{ADaFGrad} \label{alg_1}
\begin{algorithmic}[1]
\ALGORITHMHEADING{DATA:} Sequence of training datasets $\mathcal{D}^{\text{train}}=\{D^{\text{train}}_t\}^{N_t}_{t=1}$, each $D^{\text{train}}_t$ has $m$ cancer subtypes.
\ALGORITHMHEADING{INITIALIZE:} Slide buffer $\mathcal{B}_r$, text-based prototype buffer $\mathcal{B}_\mathcal{P} = \{\mathcal{B}^{cls}_\mathcal{P},\mathcal{B}^{neg}_\mathcal{P}\}$
\ALGORITHMHEADING{TRAINING PROCEDURE:} \; 
\For {$D^{\text{train}}_t$ in $\mathcal{D}^{\text{train}}$}
\State{\textcolor{blue}{Mean embeddings for $4 \times T$ descriptions for each class:}}
\State $p^{cls}_j= \frac{1}{4T}\sum^{4T}_{k=1} f^{\text{TITAN}}_{\text{text}}(cls^{j,t}_k)$, $\quad$ $\mathcal{P}^{cls}_t = \{p^{cls}_i\}^{C_t}_{i=1}$
\State{\textcolor{blue}{Mean embedding over $T$ descriptions for each negative information:}}
\State $p^{neg}_j= \frac{1}{T}\sum^{T}_{k=1} f^{\text{TITAN}}_{\text{text}}(neg^t_k)$, $\quad$ $\mathcal{P}^{neg}_t = \{p^{neg}_i\}^l_{i=1}$
\State{\textcolor{blue}{Accumulating to the prototype buffer:}}
\State $\mathcal{B}^{cls}_{\mathcal{P}} \leftarrow \mathcal{B}^{cls}_{\mathcal{P}} \cup \mathcal{P}^{cls}_t, \quad \mathcal{B}^{neg}_{\mathcal{P}} \leftarrow \mathcal{B}^{neg}_{\mathcal{P}} \cup \mathcal{P}^{neg}_t.$
\For {$e$ in $E$} \Comment{\textcolor{teal}{$E$ is the number of training epochs}}
\For {$s_i$ in $D^{\text{train}}_t$} \Comment{\textcolor{teal}{For each slide in the dataset}}
\State {\textcolor{blue}{Given region and patch embeddings $\mathcal{R}_i, \mathcal{X}_i$ for WSI $s_i$:}}
\State {$\mathbf{z}_i = f_{\mathcal{A}}^{(i)}(\mathcal{R}_i, \mathcal{X}_i)$} \Comment{\textcolor{teal}{Set of region features from last layer}}
\State Compute $\mathcal{L}_{OLVA}(\mathbf{z}_i,\mathcal{B}_{\mathcal{P}})$ \Comment{\textcolor{teal}{See Eq. \ref{eq:ovla}}}
\State{\textcolor{blue}{Compute logits for prediction:}}
\State $\hat{p}^{(i)}_i=f_{head}^{(i)}(\mathbf{z}_i, \theta^{(i)}_{head})$
\State Compute $\mathcal{L}_{CE}(\hat{p}^{(i)}_i,y_i)$ \Comment{\textcolor{teal}{Cross-entropy loss}}
\State \textcolor{blue}{Store gradient of $j$-th logit and corresponding slide:}
\State $\mathcal{B}_r \leftarrow \mathcal{B}_r \cup^{*} \left\{\left(s_i, \frac{\partial \hat{p}^{(i)}_{i,j}}{\partial \theta^{(i)}_{head}}\right)\right\}$
\State Retrieve $s_r$ and $\frac{\partial \hat{p}^{(r)}_{r,j}}{\partial \theta^{(r)}_{head}}$ from $\mathcal{B}_r$, $r < i$.
\State \textcolor{blue}{Perform replay:}
\State $\hat{p}^{(i)}_r=f_{head}^{(i)}\left(f^{(i)}_{\mathcal{A}}(\mathcal{R}_r,\mathcal{X}_r), \theta^{(i)}_{head}\right)$
\State Compute $\mathcal{L}^r_{CE}(\hat{p}^{(i)}_r,y_r)$ \Comment{\textcolor{teal}{Cross-entropy loss for replay}}
\State \textcolor{blue}{Past-to-Present Gradient Distillation:}
\State Compute $\mathcal{L}_{PPGD}\left(\frac{\partial \hat{p}^{\textcolor{red}{(r)}}_{\textcolor{red}{r},j}}{\partial \theta^{\textcolor{red}{(r)}}_{head}}, \frac{\partial \hat{p}^{\textcolor{green}{(i)}}_{\textcolor{red}{r},j}}{\partial \theta^{\textcolor{green}{(i)}}_{head}}\right)$
\Comment{\textcolor{teal}{See Eq. \ref{eq:ppgd}}}
\State Backward $\mathcal{L}_{CE}+\mathcal{L}^r_{CE} + \mathcal{L}_{OLVA} + \mathcal{L}_{PPGD}$
\EndFor
\EndFor
\EndFor
\ALGORITHMHEADING{RETURN:} The unified model $\mathcal{F} = \{f_{\mathcal{A}},f_{head}\}$ to perform classification for $N_t$ tasks.
\end{algorithmic}
\end{algorithm}

\section{Experiments}

\subsection{Datasets} We establish a sequence of $N_t = 6$ datasets covering six organs: TCGA-BRCA (breast), TCGA-RCC (kidney), TCGA-NSCLC (lung), TCGA-ESCA (esophagus), TCGA-TGCT (testis), and TCGA-CESC (cervix uteri), all downloaded from the Genomic Data Commons (GDC) Data Portal\footnote{\url{https://portal.gdc.cancer.gov/}}. To better illustrate the size of each dataset, we provide Fig.~\ref{fig:dataset}, which presents the number of cancer subtypes and the number of WSIs per subtype for each dataset. Each dataset is split into $10$ folds, each comprising a train--validation--test split. For TCGA-BRCA, TCGA-RCC and TCGA-NSCLC, we use the $0.9:0.1:0.1$ train-validation-test split ratios. For TCGA-ESCA, TCGA-TGCT, and TCGA-CESC, we use a $0.48:0.12:0.4$ split ratio. All experiments are run on 10 folds to ensure stability.

\begin{figure}[http]
\centerline{\includegraphics[width=0.4\textwidth]{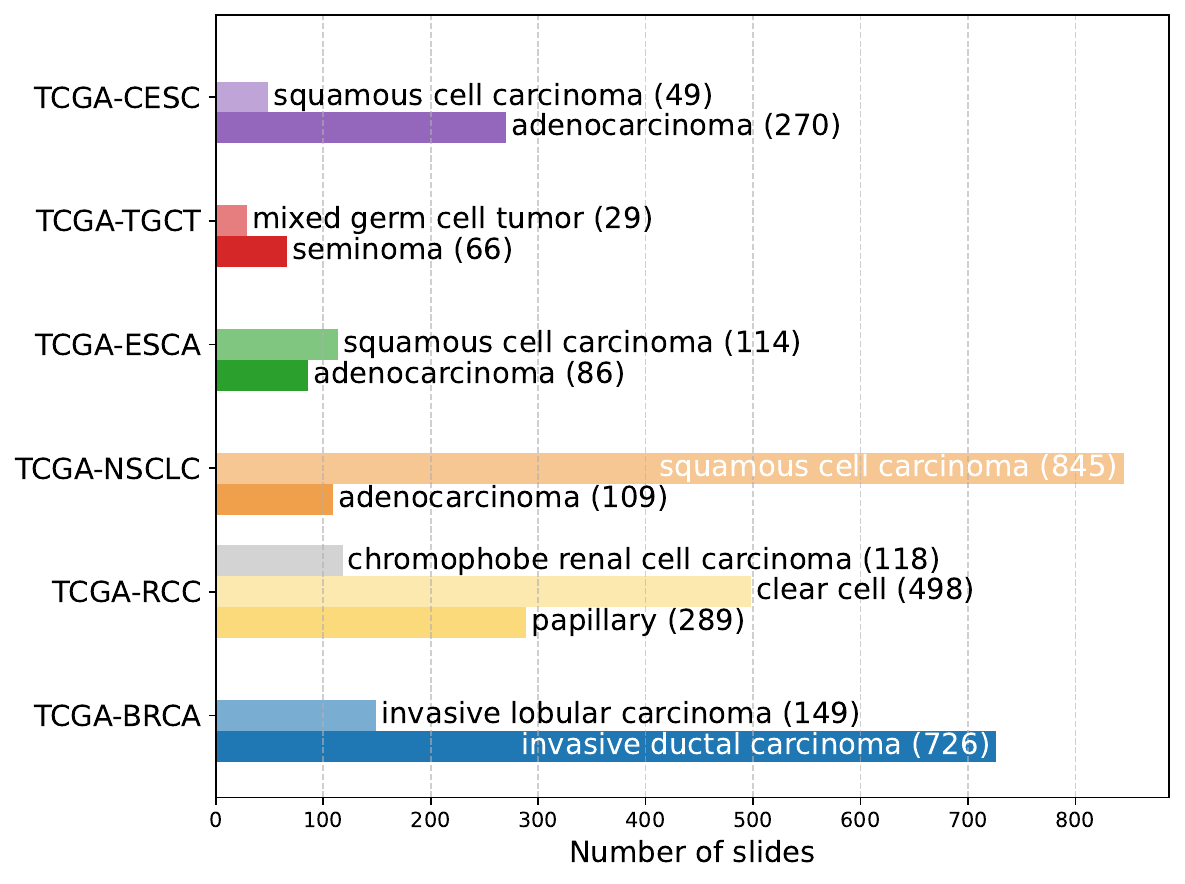}}
\caption{Distribution of the sequence of six TCGA datasets.}
\label{fig:dataset} 
\end{figure}

\subsection{Implemental Details}

All models are trained sequentially on six TCGA datasets using the same random seed to ensure stable comparisons. For the backbone $f_{\mathcal{A}}$ used to extract slide embeddings in all continual learning methods, we employ HIT \cite{huang2023conslide}, which is designed to aggregate features from the patch to region level. Regions are tiled at $10\times$ magnification into $1024 \times 1024$ pixel areas and then each region is cropped into $4 \times 4$ patches of $256 \times 256$ pixels. Both patch and region features are obtained using $f^{\mathrm{TITAN}}_{\text{vis}}$, with $C_{\text{vis}}=768$. For embedding dimension in HIT, we use $d_{\text{model}} = 384$. For all models, we train for $E = 10$ epochs per task. For ADaFGrad, only $E = 2$ epochs are needed to facilitate effective lifelong learning, demonstrating the effectiveness of our method compared to others. In the ablation study, we show that the baseline, which is trained with only $E = 2$ epochs, cannot achieve results as good as ADaFGrad. For the loss function hyperparameters, we use $\alpha=0.01$, $\beta=0.001$, $\gamma=0.2$, and $\tau=0.1$.

\begin{table*}[h]
\centering
\resizebox{0.9\textwidth}{!}{\begin{tabular}{l|c|ccc|ccc}
\toprule
\textbf{Method} & \textbf{Buffer size |$\mathcal{B}_r$|} & \textbf{ACC $\uparrow$}    & \textbf{Masked ACC $\uparrow$}  & \textbf{AUC $\uparrow$} & \textbf{FWT $\uparrow$}    & \textbf{BWT $\uparrow$}   & \textbf{FGT $\downarrow$} \\ \midrule
\multicolumn{8}{l}{\textit{Joint Training, Naive Fine-tuning and Regularization-based method}}        \\ \midrule
Joint Training (UB) & & \textcolor{gray}{84.438 ($\pm$2.096)} & \textcolor{gray}{92.466 ($\pm$1.058)} & \textcolor{gray}{0.966 ($\pm$0.013)} & -- & -- & -- \\
Fine-tuning (LB) & & \textcolor{gray}{27.480 ($\pm$2.732)} & \textcolor{gray}{85.774 ($\pm$2.602)} & \textcolor{gray}{0.947 ($\pm$0.015)} & \textcolor{gray}{-3.187 ($\pm$2.722)} & \textcolor{gray}{-8.035 ($\pm$3.795)} & \textcolor{gray}{8.794 ($\pm$3.053)} \\
EWC \cite{kirkpatrick2017overcoming} & \multirow{-4}{*}{0 WSIs} & 43.522 ($\pm$6.765) & 89.244 ($\pm$1.902) & 0.954 ($\pm$0.016) & 0.810 ($\pm$5.004) & -3.617 ($\pm$3.101) & 4.649 ($\pm$2.558) \\ \midrule
\multicolumn{8}{l}{\textit{Rehearsal-based Methods}}    \\ \midrule
GDumb \cite{gdumb} &  & 21.662 ($\pm$8.443) & 51.671 ($\pm$6.426) & 0.659 ($\pm$0.081) & 0.000 ($\pm$0.000) & 11.792 ($\pm$8.533) & 4.486 ($\pm$3.529) \\
{ER-ACE \cite{erace}} &  & 51.297 ($\pm$7.962) & 77.084 ($\pm$3.211) & 0.882 ($\pm$0.051) & 6.553 ($\pm$5.268) & -0.320 ($\pm$1.211) & 5.095 ($\pm$2.429) \\
{AGEM \cite{agem}}  &  & 47.606 ($\pm$4.657) & 85.850 ($\pm$2.543) & 0.915 ($\pm$0.028) & -1.344 ($\pm$3.341) & -7.714 ($\pm$3.978) & 8.167 ($\pm$3.836) \\
{DER++ \cite{derpp}}    &  & {{56.619 ($\pm$4.027)}} & {{89.571 ($\pm$1.051)}} & {{0.948 ($\pm$1.901)}} & {{-0.408 ($\pm$5.129)}} & {{-3.627 ($\pm$1.182)}}& {{4.266 ($\pm$1.089)}} \\
{ConSlide \cite{huang2023conslide}} &  & {{64.226 ($\pm$4.282)}} & {{89.318 ($\pm$2.349)}} & {{0.954 ($\pm$1.679)}} & {{-5.915 ($\pm$2.072)}} & {{0.196 ($\pm$1.745)}} & {{4.849 ($\pm$2.527)}} \\
{\textbf{ADaFGrad (ours)}} & \multirow{-6}{*}{$\approx 10$ WSIs} & \cellcolor[HTML]{FFFACD}{{\textbf{67.247 ($\pm$3.108)}}} & \cellcolor[HTML]{FFFACD}{\textbf{{90.876 ($\pm$2.048)}}} & \cellcolor[HTML]{FFFACD}{{\textbf{0.957 ($\pm$0.015)}}} & \cellcolor[HTML]{FFFACD}{{\textbf{-0.359 ($\pm$2.510)}}} & \cellcolor[HTML]{FFFACD}{{\textbf{-1.589 ($\pm$1.693)}}} & \cellcolor[HTML]{FFFACD}{{\textbf{2.772 ($\pm$1.633)}}} \\ \midrule
GDumb \cite{gdumb} &  & 28.308 ($\pm$7.657) & 64.438 ($\pm$4.084) & 0.615 ($\pm$0.073) & 0.000 ($\pm$0.000) & 20.331 ($\pm$5.479) & 2.676 ($\pm$1.507) \\
{ER-ACE \cite{erace}}  &  & 68.042 ($\pm$4.296) & 84.736 ($\pm$3.236) & 0.937 ($\pm$0.021) & 7.158 ($\pm$4.948) & 0.088 ($\pm$3.317) & 4.251 ($\pm$2.606) \\
{AGEM \cite{agem}}     &  & 47.606 ($\pm$4.657) & 85.850 ($\pm$2.543) & 0.915 ($\pm$0.028) & -1.344 ($\pm$3.341) & -7.714 ($\pm$3.978) & 8.167 ($\pm$3.836) \\
{DER++ \cite{derpp}}    &  & {{58.998 ($\pm$1.219)}} & {{90.604 ($\pm$1.472)}} & {{0.956 ($\pm$0.175)}} & {{6.027 ($\pm$3.589)}}  & {{-2.368 ($\pm$1.889)}}& {{3.199 ($\pm$1.640)}} \\
{ConSlide \cite{huang2023conslide}} &  & {{65.673 ($\pm$1.780)}} & {{90.255 ($\pm$1.334)}} & {{0.953 ($\pm$1.570)}} & {{-6.514 ($\pm$2.409)}} & {{-2.930 ($\pm$1.506)}}& {{4.032 ($\pm$1.248)}} \\
{\textbf{ADaFGrad (ours)}} & \multirow{-6}{*}{$\approx 30$ WSIs} & \cellcolor[HTML]{FFFACD}{{\textbf{70.741 ($\pm$3.569)}}} & \cellcolor[HTML]{FFFACD}{{\textbf{91.235 ($\pm$1.641)}}} & \cellcolor[HTML]{FFFACD}{{\textbf{0.957 ($\pm$0.020)}}} & \cellcolor[HTML]{FFFACD}{{\textbf{-0.563 ($\pm$3.467)}}} & \cellcolor[HTML]{FFFACD}{{\textbf{-1.217 ($\pm$1.805)}}} & \cellcolor[HTML]{FFFACD}{{\textbf{2.517 ($\pm$1.125)}}} \\ \bottomrule
\end{tabular}}
\caption{Experimental results comparing naive fine-tuning, regularization-based methods, and rehearsal-based methods on a sequence of six TCGA datasets.}
\label{tab:main}
\end{table*}

\subsection{Evaluation Metrics}

We evaluate continual learning performance using seven metrics:  
(1) Accuracy (ACC),  
(2) Masked Accuracy (Masked~ACC),  
(3) Area Under the Curve (AUC),  
(4) Mean Accuracy (mACC),  
(5) Backward Transfer (BWT),  
(6) Forward Transfer (FWT), and  
(7) Forgetting (FGT).  Let \(N_t\) be the total number of tasks, and let \(ACC_{t}(i)\) denote the accuracy on task \(i\) after training up through task \(t\).

\medskip
\noindent\textbf{Accuracy (ACC).}  
Under the \textbf{CLASS-IL} scenario, the model $\mathcal{F}$ predicts a logit vector for $i$-th WSI in the test dataset: \(\hat{p}_i\in\mathcal{C}\). The cancer subtyping class is predicted by
\[
\hat{y} = \argmax_{j}\,\hat{p}_{i,j},
\]
and we report overall classification accuracy.

\noindent\textbf{Masked Accuracy (Masked~ACC).}  
Under \textbf{TASK-IL}, we know which task \(d\) each sample belongs to, so we mask out all other logits. If the $t$-th task’s logits occupy indices 
$
s_t = \sum_{i=1}^{t-1}C_i,\quad e_t = \sum_{i=1}^{t}C_i
$, we compute
\[
\hat{y} = \argmax_{j\in\{s_d,\dots,e_d\}}\,\hat{p}_{i,j}
\]
and measure accuracy only over these masked logits.

\noindent\textbf{Area Under the Curve (AUC).} Because the datasets in the sequence have different numbers of classes (for example, TCGA-RCC has three subtyping classes), we compute the AUC for each class in each test dataset and then average these values across all classes.

\noindent\textbf{Mean Accuracy (mACC).}  
The mean accuracy is the average over all intermediate mean-task accuracies:
\begin{equation}
\mathrm{mACC}
=\frac{1}{T}\sum_{t=1}^{N_t}\Bigl(\frac{1}{t}\sum_{i=1}^{t}ACC_{t}(i)\Bigr).
\label{eq:mean_acc}
\end{equation}

\noindent\textbf{Backward Transfer (BWT).}  
Backward Transfer measures how learning new tasks affects performance on old tasks:
\begin{equation}
\mathrm{BWT}
=\frac{1}{N_t-1}
\sum_{t=1}^{N_t-1}\bigl(ACC_{N_t}(t)-ACC_{t}(t)\bigr).
\end{equation}

\noindent\textbf{Forward Transfer (FWT).}  
Forward Transfer quantifies how training on earlier tasks improves performance on future tasks, relative to a random baseline \(ACC^{\mathrm{rand}}(t)\):
\begin{equation}
\mathrm{FWT}
=\frac{1}{N_t-1}
\sum_{t=2}^{N_t}\bigl(ACC_{t-1}(t)-ACC^{\mathrm{rand}}(t)\bigr).
\end{equation}

\noindent\textbf{Forgetting (FGT).}  
Forgetting measures the drop from each task’s best achieved accuracy to its final accuracy after all training:
\begin{equation}
\mathrm{FGT}
=\frac{1}{T}\sum_{t=1}^{T}
\Bigl(\max_{d\in\{t,\dots,T\}}ACC_{d}(t)-ACC_{T}(t)\Bigr).
\end{equation}

\subsection{Experiments}

\begin{figure}[!h]
\centerline{\includegraphics[width=0.45\textwidth]{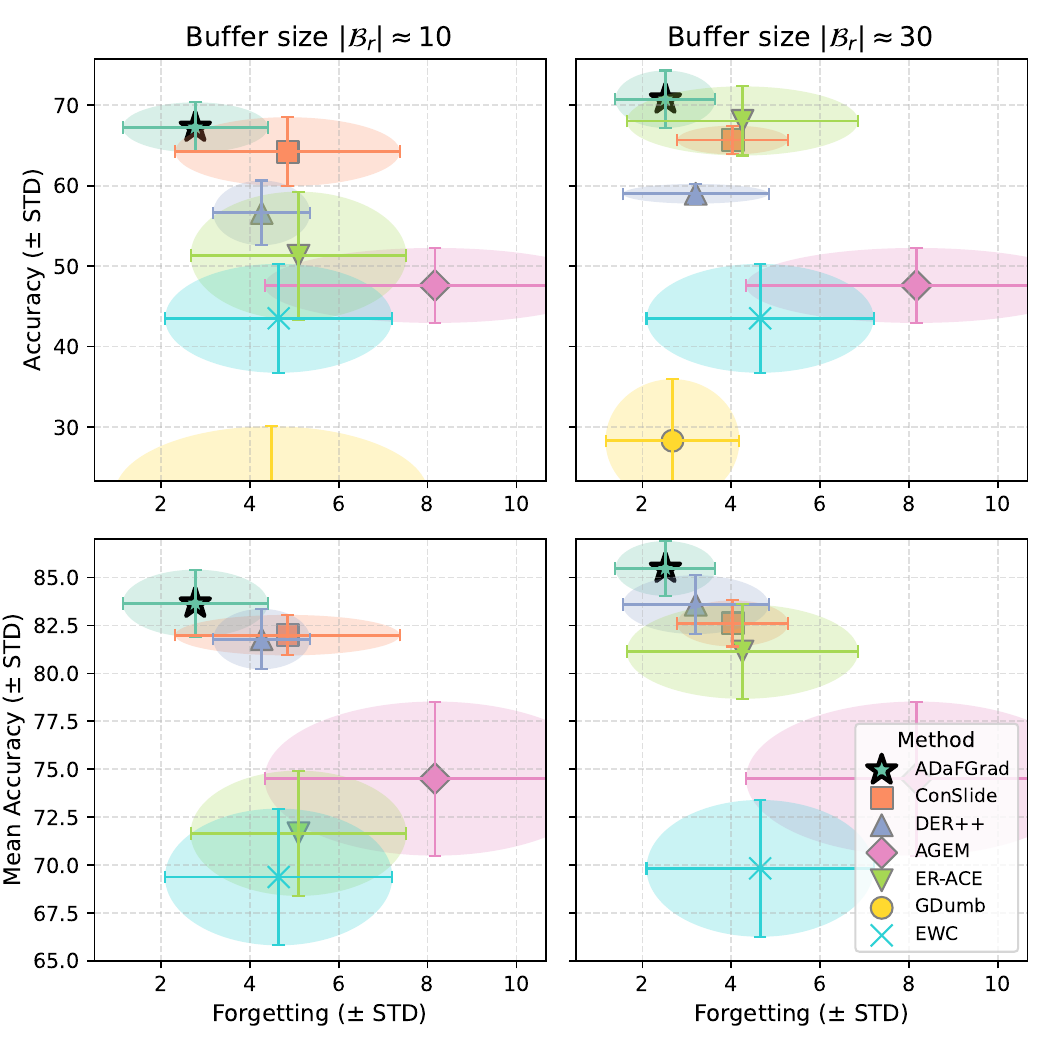}}
\caption{Accuracy–Forgetting trade-off. The first row shows the ACC–FGT trade-off, and the second row shows the mACC–FGT trade-off.}
\label{fig:trade-off} 
\end{figure}

\begin{figure*}[http]
\centerline{\includegraphics[width=1\textwidth]{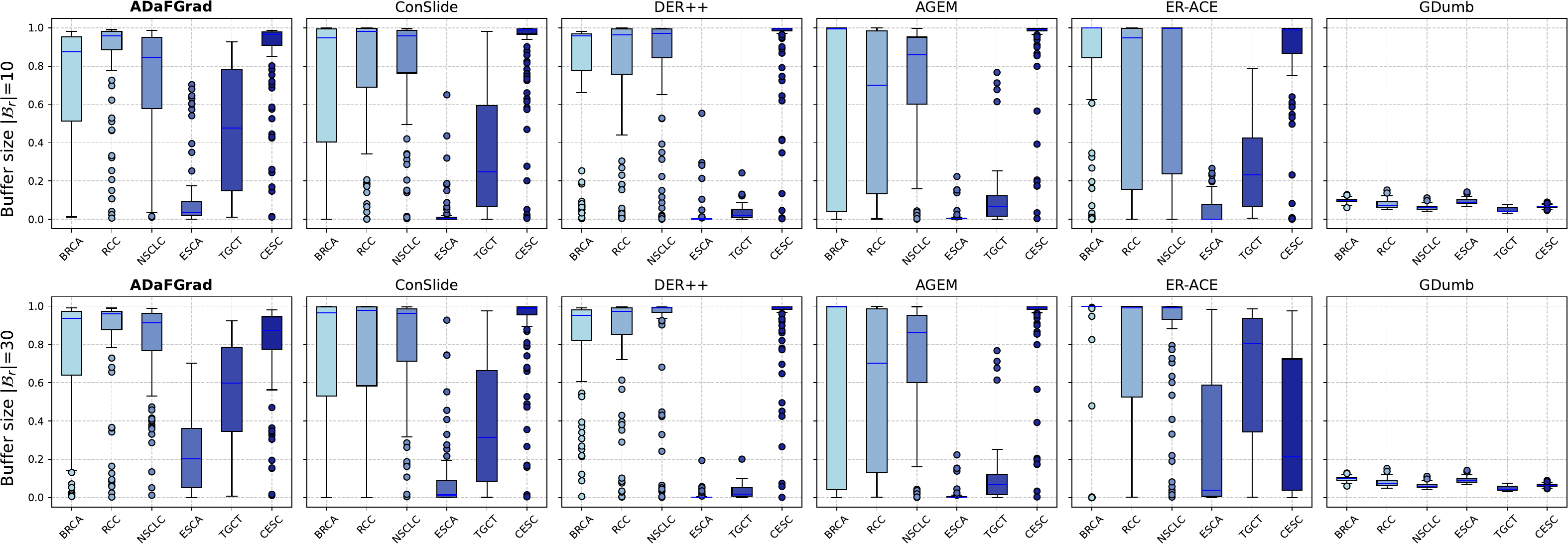}}
\caption{Confidence scores for target cancer subtype labels after training on the final tasks of ADaFGrad and all other models.}
\label{fig:logit} 
\end{figure*}

\begin{figure*}[http]
\centerline{\includegraphics[width=0.88\textwidth]{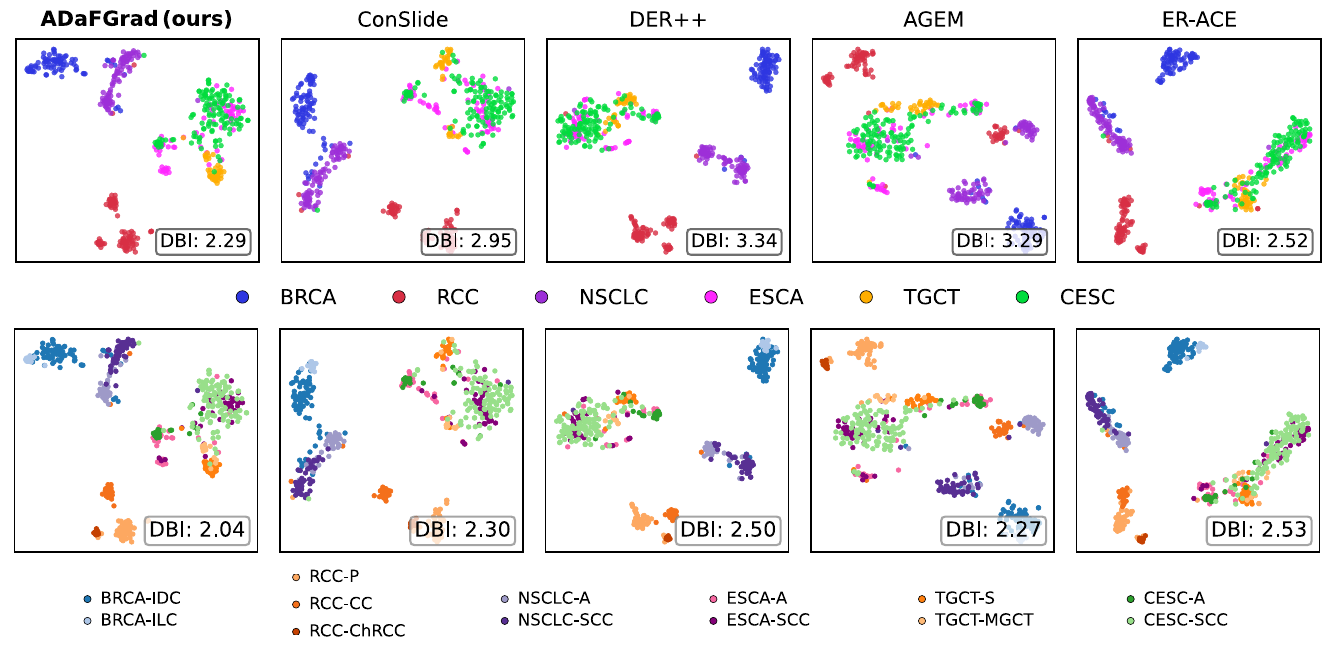}}
\caption{t-SNE \cite{van2008visualizing} visualization of slide embeddings from all continual learning methods, clustered by task (first row) and by class (second row) spaces. Abbreviations: IDC, invasive ductal carcinoma; ILC, invasive lobular carcinoma; P, papillary carcinoma; CC, clear cell carcinoma; ChRCC, chromophobe renal cell carcinoma; A, adenocarcinoma; SCC, squamous cell carcinoma; S, seminoma; MGCT, mixed germ cell tumor.}
\label{fig:tsne} 
\end{figure*}

\subsubsection{Comparative Models} \noindent For comparison, we include EWC as the regularization-based model, and GDumb, ER-ACE, AGEM, DER++, and ConSlide as rehearsal-based models. It is worth noting that all continual-learning results are upper-bounded by \textbf{Joint Training} (UB), in which the model is trained on all datasets simultaneously, and lower-bounded by \textbf{Fine-tuning} (LB), in which \(\mathcal{F}\) is trained sequentially task by task without applying any continual-learning techniques. For all rehearsal-based models, we evaluate ADaFGrad against the others under two buffer sizes: $|\mathcal{B}_r| = 10$ and $|\mathcal{B}_r| = 30$. As mentioned, all methods use HIT \cite{huang2023conslide} as the backbone network. For ConSlide and ADaFGrad, the BuRo buffering strategy is applied to diversify the buffer by storing tissue regions rather than entire WSIs. Therefore, $|\mathcal{B}_r| = 1100 \approx 10$ and $|\mathcal{B}_r| = 6600 \approx 30$ slides are used for these methods.

\subsubsection{Main Results}

\noindent \textbf{ACC, Masked ACC and AUC.} Main experimental results are reported in Tab.~\ref{tab:main}. First, ADaFGrad surpasses all other models in ACC, Masked ACC, and AUC regardless of buffer size. Specifically, under $|\mathcal{B}_r|=10$, ADaFGrad achieves improvements of $>3.021\%$ in ACC, $>1.558\%$ in Masked ACC, and $>0.003$ in AUC. With $|\mathcal{B}_r|=30$, the gap between ADaFGrad and the WSI‐specific model ConSlide increases to $+5.068\%$ ACC, $+0.980\%$ Masked ACC, and $+0.004$ in AUC. Additionally, several observations emerge: (1) all continual‐learning methods progressively approach the performance of Joint Training for WSI analysis; (2) GDumb is not effective on WSIs; and (3) the CLASS‐IL scenario is much more challenging than TASK‐IL for WSI analysis. For example, at $|\mathcal{B}_r|=30$, although DER++ and ConSlide remain competitive with ADaFGrad in Masked ACC (only $-0.631\%$ and $-0.980\%$, respectively), their ACC margins are significant at $-11.743\%$ and $-5.068\%$, respectively.

\noindent \textbf{FWT, BWT and FGT.} These metrics assess method stability as the number of tasks increases. It is worth noting, however, that although higher BWT and FWT are better, they may coincide with low overall accuracy. For instance, ER-ACE achieves high FWT, and GDumb achieves high BWT, but their ACC and Masked ACC scores are not as good as ADaFGrad’s. FGT exhibits superior stability; it measures the gap between a task’s best performance and its performance after training on the final task. Therefore, we prioritize a good balance between FGT and ACC in the CLASS-IL scenario. We visualize this trade-off in Fig. \ref{fig:trade-off}. Because lower FGT and higher ACC are preferable, an effective method should lie as close as possible to the top-left corner. ADaFGrad clearly achieves the best stability in the ACC-FGT trade-off, and its standard deviations across folds are among the most consistent.

\noindent \textbf{Mean Accuracies.} The mean accuracy over accumulated dataset sequences also indicates method quality and stability. This metric is defined in Eq.~\ref{eq:mean_acc}. We apply it to CLASS-IL (mACC), TASK-IL (mean Masked ACC), and AUC (mAUC). The results are reported in Tab.~\ref{tab:mean}. Consistent with the main results in Tab.~\ref{tab:main}, ADaFGrad remains the best performer, with mACC improvements of $>1.659\%$ and $>2.867\%$ for $|\mathcal{B}_r|=10$ and $|\mathcal{B}_r|=30$, respectively. Its mean Masked ACC and mAUC are also stable and competitive with all other methods. The mACC-FGT trade-off is likewise the best, as shown in Fig.~\ref{fig:trade-off}.

\begin{table}[]
\centering
\resizebox{0.49\textwidth}{!}{\begin{tabular}{l|c|ccc}
\toprule
\textbf{Method}  & \textbf{Buffer size} & \textbf{mACC $\uparrow$} & \textbf{mean Masked ACC $\uparrow$} & \textbf{mAUC $\uparrow$} \\ \midrule
Fine-tuning (LB) & \multirow{3}{*}{$0$} & 59.779 ($\pm$1.954) & 86.065 ($\pm$2.783) & 0.958 ($\pm$0.016) \\
EWC \cite{kirkpatrick2017overcoming}  &  & 69.834 ($\pm$3.566) & 89.575 ($\pm$2.730) & 0.961 ($\pm$0.017) \\ \midrule
GDumb \cite{gdumb} & \multirow{6}{*}{$\approx10$}  & 9.207 ($\pm$2.070) & 44.186 ($\pm$2.604) & 0.570 ($\pm$0.022) \\
ER-ACE \cite{erace} &  & 71.663 ($\pm$3.265) & 82.447 ($\pm$2.726) & 0.938 ($\pm$0.022) \\
AGEM \cite{agem} &  & 74.514 ($\pm$4.016) & 87.278 ($\pm$3.696) & 0.927 ($\pm$0.029) \\
DER++ \cite{derpp} &  & 81.786 ($\pm$1.575) & 90.850 ($\pm$1.558) & 0.956 ($\pm$0.019) \\
ConSlide \cite{huang2023conslide} &  & 81.992 ($\pm$1.055) & 90.572 ($\pm$0.977) & 0.964 ($\pm$0.016) \\
ADaFGrad (ours) &  & \cellcolor[HTML]{FFFACD}83.651 ($\pm$1.755) & \cellcolor[HTML]{FFFACD}90.877 ($\pm$1.657) & \cellcolor[HTML]{FFFACD}0.962 ($\pm$0.018) \\ \midrule
GDumb \cite{gdumb} & \multirow{6}{*}{$\approx30$}  & 10.313 ($\pm$2.401) & 46.313 ($\pm$2.689) & 0.563 ($\pm$0.024) \\
ER-ACE \cite{erace} &  & 81.138 ($\pm$2.464) & 87.669 ($\pm$2.284) & 0.951 ($\pm$0.022) \\
AGEM \cite{agem} &  & 74.514 ($\pm$4.016) & 87.278 ($\pm$3.696) & 0.927 ($\pm$0.029) \\
DER++ \cite{derpp} &  & 83.580 ($\pm$1.532) & 91.637 ($\pm$1.476) & 0.959 ($\pm$0.018) \\
ConSlide \cite{huang2023conslide} &  & 82.602 ($\pm$1.201) & 91.092 ($\pm$1.067) & 0.964 ($\pm$0.016) \\
ADaFGrad (ours) &  & \cellcolor[HTML]{FFFACD}85.469 ($\pm$1.430) & \cellcolor[HTML]{FFFACD}91.524 ($\pm$1.361) & \cellcolor[HTML]{FFFACD}0.962 ($\pm$0.020) \\ \bottomrule
\end{tabular}}
\caption{Experimental results on comparative methods in terms of mean Accuracy, mean Masked Accuracy, and mean AUC.}
\label{tab:mean}
\end{table}

\begin{table}[h]
\centering
\resizebox{0.49\textwidth}{!}{\begin{tabular}{ccc|ccc}
\toprule
$\mathcal{L}^r_{CE}$ & $\mathcal{L}_{OVLA}$ & $\mathcal{L}_{PPGD}$ & \textbf{ACC $\uparrow$} & \textbf{Masked ACC $\uparrow$} & \textbf{mACC $\uparrow$} \\ \midrule
\checkmark & \xmark & \xmark & 30.657 ($\pm$3.594) & 85.818 ($\pm$2.734) & 57.585 ($\pm$2.012) \\
\checkmark & \checkmark & \xmark & 68.040 ($\pm$3.222) & 90.651 ($\pm$1.585) & 82.439 ($\pm$1.837) \\
\checkmark & \checkmark  & \checkmark & \cellcolor[HTML]{FFFACD}\textbf{70.741 ($\pm$3.569)} & \cellcolor[HTML]{FFFACD}\textbf{91.235 ($\pm$1.641)} & \cellcolor[HTML]{FFFACD}\textbf{85.469 ($\pm$1.430)} \\ \midrule
$\mathcal{L}^r_{CE}$ & $\mathcal{L}_{OVLA}$ & $\mathcal{L}_{PPGD}$ & \textbf{FWT $\uparrow$} & \textbf{BWT $\uparrow$} & \textbf{FGT $\downarrow$} \\ \midrule
\checkmark & \xmark & \xmark & 2.675 ($\pm$1.789) & -7.535 ($\pm$3.061) & 8.702 ($\pm$2.480) \\
\checkmark & \checkmark & \xmark & -0.570 ($\pm$3.054) & -1.989 ($\pm$1.862) & 2.853 ($\pm$1.444) \\
\checkmark & \checkmark & \checkmark & \cellcolor[HTML]{FFFACD}\textbf{-0.563 ($\pm$3.467)} & \cellcolor[HTML]{FFFACD}\textbf{-1.217 ($\pm$1.805)} & \cellcolor[HTML]{FFFACD}\textbf{2.517 ($\pm$1.125)} \\ \bottomrule
\end{tabular}}
\caption{Ablation study results investigating the contributions of $\mathcal{L}_{OVLA}$ and $\mathcal{L}_{PPGD}$. The baseline uses HIT \cite{huang2023conslide} with only $\mathcal{L}^r_{CE}$ applied to slides sampled from $\mathcal{B}_r$.}
\label{tab:ablation}
\end{table}

\subsection{Ablation Analysis}

To further demonstrate the effectiveness of ADaFGrad, we conduct an ablation study that verifies the contributions of two loss modules: $\mathcal{L}_{OVLA}$ and $\mathcal{L}_{PPGD}$. The baseline, which does not apply either of these losses, uses the model $\mathcal{F}$ with HIT as its backbone and performs replay supervised by $\mathcal{L}^r_{CE}$ on slides sampled from the buffer. \textit{Notably, both the baseline and full ADaFGrad are trained for only two epochs per task.} The results are reported in Tab. \ref{tab:ablation}. By adding our $\mathcal{L}_{OVLA}$ to align the bag of region features with their corresponding class prototypes, all metrics show significant increases, especially a $+37.383\%$ improvement in ACC. With only a few epochs per task, the baseline easily collapses as the number of tasks increases. Applying $\mathcal{L}_{PPGD}$ yields slight gains in ACC; however, in mACC the performance improves substantially by $+3.030\%$. Furthermore, the stability metrics also increase for FWT, BWT, and FGT by $+0.007\%$, $+0.772\%$, and $-0.336\%$, respectively. The standard deviation of FGT is more stable ($\pm1.125$ compared to $\pm1.444$), indicating better stability when performing past-to-present gradient distillation.

\subsection{Qualitative Results} To further demonstrate ADaFGrad’s effectiveness qualitatively, we examine two aspects: \textit{(1) the confidence score of the target label in the predicted logits}, and \textit{(2) the quality of embeddings in both task space and class space}.

\noindent\textbf{Confidence Score Study.}  We visualize a box plot of the confidence scores produced by the model $\mathcal{F}$ for the target class after training on all tasks (Fig.~\ref{fig:logit}). Each box corresponds to all samples in the $t$-th test dataset $D^{\text{text}}_t$. \textit{A stable model should assign high confidence to the correct class.} We compare two buffer sizes, $|\mathcal{B}_r|=10$ and $|\mathcal{B}_r|=30$. Under $|\mathcal{B}_r|=10$, all models tend to have low confidence on the TCGA-ESCA and TCGA-TGCT test sets. ADaFGrad achieves the highest median confidence on TCGA-TGCT ($\approx 0.42$), whereas the other models remain below $0.3$. For TCGA-ESCA, aside from GDumb (which appears to guess at random), ADaFGrad’s median confidence exceeds $0.1$, while the others have medians near zero. Under $|\mathcal{B}_r|=30$, ADaFGrad maintains the highest median confidence on TCGA-ESCA ($\approx 0.2$), compared to near-zero for the other methods. On TCGA-TGCT, ADaFGrad’s median is $\approx 0.6$, although ER-ACE achieves a slightly higher value. However, for TCGA-CESC, ADaFGrad outperforms ER-ACE: ER-ACE’s median drops to $\approx 0.2$ compared to $\approx 0.9$ for ADaFGrad. Across all other test sets, ADaFGrad demonstrates the greatest stability, consistently outperforming the other methods.

\noindent\textbf{t-SNE Visualization.} For all methods, we extract slide embeddings produced by $\mathcal{F}$ with embedding dimension $\mathrm{dim}=384$ and project them into $2\mathrm{D}$ space using t-SNE \cite{van2008visualizing}. We then color the resulting scatter points by task and by class, visualized in Fig. \ref{fig:tsne}. A good model should produce slide embeddings whose $2\mathrm{D}$ projections cluster well according to their task and class labels. Qualitative t-SNE results may show ambiguous clustering across methods. Hence, we additionally use the Davies–Bouldin index (DBI) \cite{davies2009cluster} as a quantitative measure to evaluate clustering quality, where lower values indicate better clusters. Regardless of whether we cluster by task or by class, ADaFGrad achieves the lowest DBI scores of $2.29$ and $2.04$, respectively. Qualitatively, for task clustering, all methods exhibit ambiguous embeddings for TCGA-ESCA, TCGA-TGCT, and TCGA-CESC. This observation aligns with Fig.~\ref{fig:logit}, where confidence scores for TCGA-ESCA and TCGA-TGCT are low, but those for TCGA-CESC are high. It is possible that all methods are biased toward TCGA-CESC, which is the final task. Even so, ADaFGrad still clusters TCGA-CESC and TCGA-TGCT effectively, although TCGA-ESCA remains highly overlapped with those two.  

\subsection{Discussion}

\noindent \includegraphics[height=2ex]{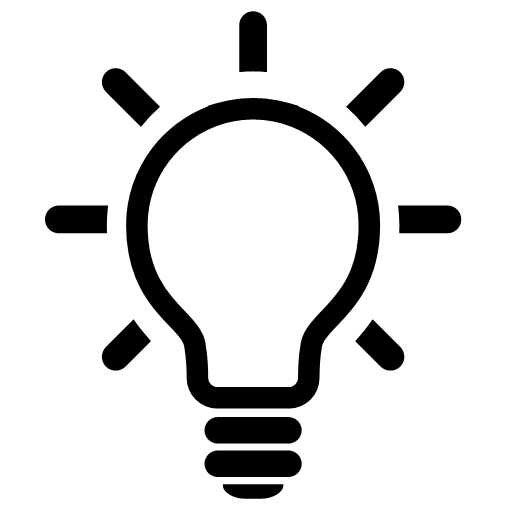}\textbf{ADaFGrad achieves strong performance with minimal epochs.} ADaFGrad demonstrates superior performance compared to other continual learning models while using only two epochs per task compared to ten epochs in all scenarios: CLASS-IL, TASK-IL, and buffer sizes $|\mathcal{B}_r|=10$ and $|\mathcal{B}_r|=30$. Ablation results indicate that most of this improvement comes from $\mathcal{L}_{OVLA}$, which performs contrastive learning between tissue region embeddings and text-based prototypes, significantly shortening the training process. The addition of $\mathcal{L}_{PPGD}$ further improves performance by preventing excessive model drift when new tasks are introduced, as shown in Fig.~\ref{fig:logit}. These characteristics facilitate practical deployment in clinical and hospital settings for WSI-related tasks on large volumes of slides in a distributed and continual manner. 

\noindent \includegraphics[height=2ex]{figures/light-bulb.png}\textbf{ADaFGrad outperforms zero-shot continual learning (ZeroSlide).} This success is largely enabled by the well pretrained vision–language foundation model TITAN. We therefore investigate whether a training-free zero-shot classification using TITAN alone could achieve competitive results. \citeauthor{bui2025zeroslide} \cite{bui2025zeroslide} compared zero-shot classification with TITAN in a continual learning setting (ZeroSlide) against other methods and reported surprisingly competitive performance. Nevertheless, our head-to-head comparison in Tab.~\ref{tab:zeroslide} shows that ADaFGrad still outperforms ZeroSlide. Thus, with a small buffer and only a few training epochs, ADaFGrad exceeds the accuracy of training-free approaches. Considering the trade-off between training cost and accuracy, ADaFGrad remains the most effective method.  

\noindent \includegraphics[height=2ex]{figures/light-bulb.png}\textbf{Inference incurs no additional overhead.} It is worth noting that our $\mathcal{L}_{OVLA}$ and $\mathcal{L}_{PPGD}$ introduce extra memory and computational costs during training, such as prototype buffers and gradient memory. However, neither is used during inference. Some studies likewise incur additional inference overhead; for example, CATE \cite{huang2024free} requires extra computation for text embeddings and aggregation with a bag of patch features.

Finally, we discuss the limitations of ADaFGrad. This study focuses on WSI classification tasks, specifically cancer subtyping, where predictions are discrete logits. To extend this framework to tasks such as survival analysis \cite{zhu2017wsisa,tang2019capsurv,liu2024advmil}, biochemical recurrence \cite{eminaga2024artificial}, or tissue segmentation \cite{bui2024efficient} further investigation is needed to select appropriate prototypes for $\mathcal{L}_{OVLA}$ and to design suitable gradient strategies for $\mathcal{L}_{PPGD}$.

\begin{table}[]
\centering
\caption{Comparison against training-free ZeroSlide \cite{bui2025zeroslide}.}
\resizebox{0.49\textwidth}{!}{\begin{tabular}{lccc}
\toprule
\textbf{Method} & \textbf{ACC $\uparrow$} & \textbf{Masked ACC $\uparrow$} & \textbf{mACC} $\uparrow$ \\ \midrule
ZeroSlide \cite{bui2025zeroslide} & 64.129 ($\pm$1.591)  & 89.758 ($\pm$0.984) & 82.593 ($\pm$0.017) \\
ADaFGrad ($|\mathcal{B}_r|\approx10$) & 67.247 ($\pm$3.108)  & 90.876 ($\pm$2.048) & 83.651 ($\pm$1.755) \\
ADaFGrad ($|\mathcal{B}_r|\approx30$) & \textbf{70.741 ($\pm$3.569)} & \textbf{91.235 ($\pm$1.641)} & \textbf{85.469 ($\pm$1.430)} \\
\bottomrule
\end{tabular}}
\label{tab:zeroslide}
\end{table}

\section{Conclusion}

This study introduces ADaFGrad for lifelong learning in WSI analysis. ADaFGrad comprises two key modules: Online Vision–Language Adaptation (OVLA) and Past-to-Present Gradient Distillation (PPGD). OVLA leverages a vision–language foundation model to align a slide’s bag-of-region embeddings with text-based prototype embeddings via contrastive learning, while PPGD preserves model stability by mimicking the current gradient of the target logit with respect to the classification head’s parameters, preventing significant shifts when new tasks are added sequentially. Experimental results on six TCGA datasets for cancer subtyping demonstrate ADaFGrad’s effectiveness both qualitatively and quantitatively. Future work will extend this approach to multi-task lifelong learning in WSI analysis, encompassing tasks such as cancer subtyping, cancer grading, survival analysis, and biochemical recurrence.

\printbibliography
\vspace{-4cm}
\begin{IEEEbiography}[{\includegraphics[width=1in,height=1.25in,clip,keepaspectratio]{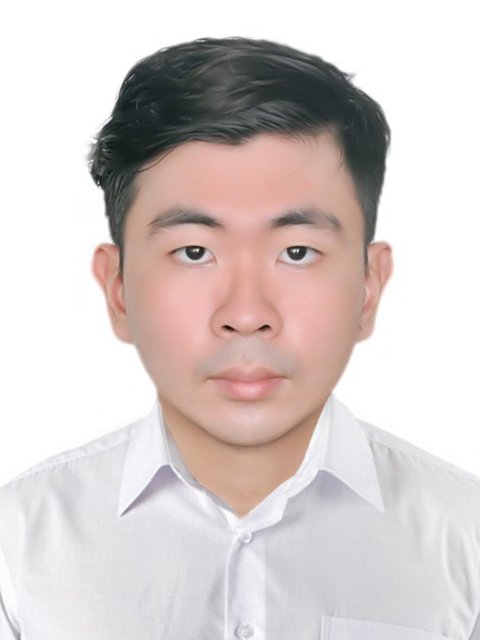}}]{Doanh C. Bui} received his B.S. degree in Computer Science from the University of Information Technology, VNU-HCM, Vietnam, in 2022, and his M.S. degree in Electrical and Computer Engineering from Korea University, Seoul, Republic of Korea, in 2025. He is currently a Ph.D. student at the Architecture Lab at Nara Institute of Science and Technology, Japan. His research interests focus on vision-language models and digital/computational pathology. 
He has authored or co-authored several papers in leading journals such as IEEE Transactions on Medical Imaging (TMI) and Computer Methods and Programs in Biomedicine (CMPB), as well as in top-tier conferences including MICCAI, WACV, AAAI, and ACCV.
\end{IEEEbiography}
\vskip -3.1\baselineskip plus -1fil
\begin{IEEEbiography}[{\includegraphics[width=1in,height=1.25in,clip,keepaspectratio]{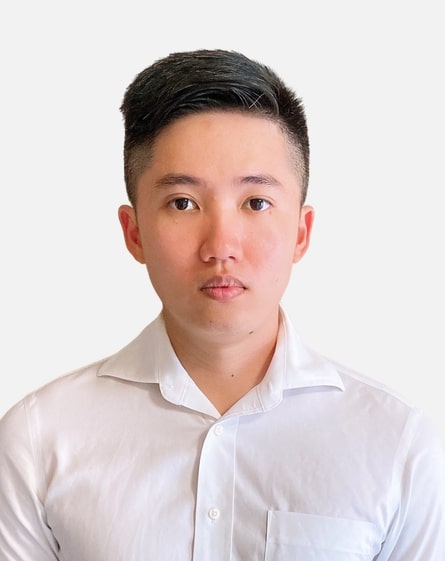}}]{Hoai Luan Pham} (Member, IEEE) received a bachelor’s degree in Computer Engineering from Vietnam National University Ho Chi Minh City—University of Information Technology (UIT), Vietnam, in 2018, and a master’s degree and Ph.D. degree in Information Science from the Nara Institute of Science and Technology (NAIST), Japan, in 2020 and 2022, respectively. Since October 2022, he has been with NAIST as an Assistant Professor and with UIT as a Visiting Lecturer. His research interests include blockchain technology, cryptography, computer architecture, circuit design, and accelerators.
\end{IEEEbiography}
\vskip -3\baselineskip plus -1fil
\begin{IEEEbiography}[{\includegraphics[width=1in,height=1.25in,clip,keepaspectratio]{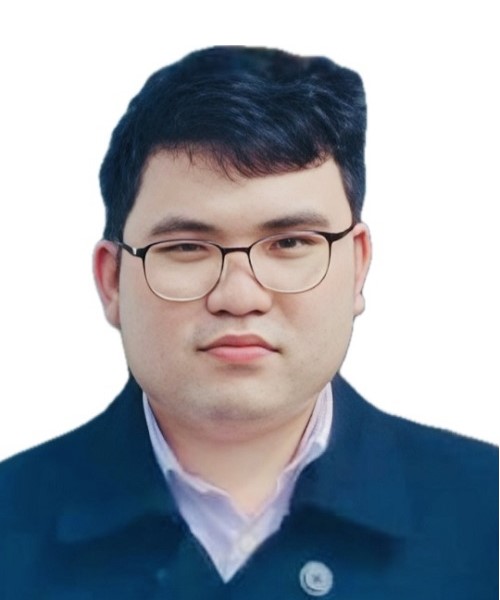}}]{Vu Trung Duong Le} (Member, IEEE) received the B.Eng. degree in Integrated Circuits and Hardware Design from the University of Information Technology (UIT), VNU-HCM, in 2020, and the M.S. and Ph.D. degrees in Information Science from the Nara Institute of Science and Technology (NAIST) in 2022 and 2024, respectively. He is currently an Assistant Professor at the Computing Architecture Laboratory, NAIST. His research interests include computing architecture, reconfigurable processors, and accelerator design for artificial intelligence, quantum emulation and cryptography.
\end{IEEEbiography}
\vskip -3.1\baselineskip plus -1fil
\begin{IEEEbiography}[{\includegraphics[width=1in,height=1.25in,clip,keepaspectratio]{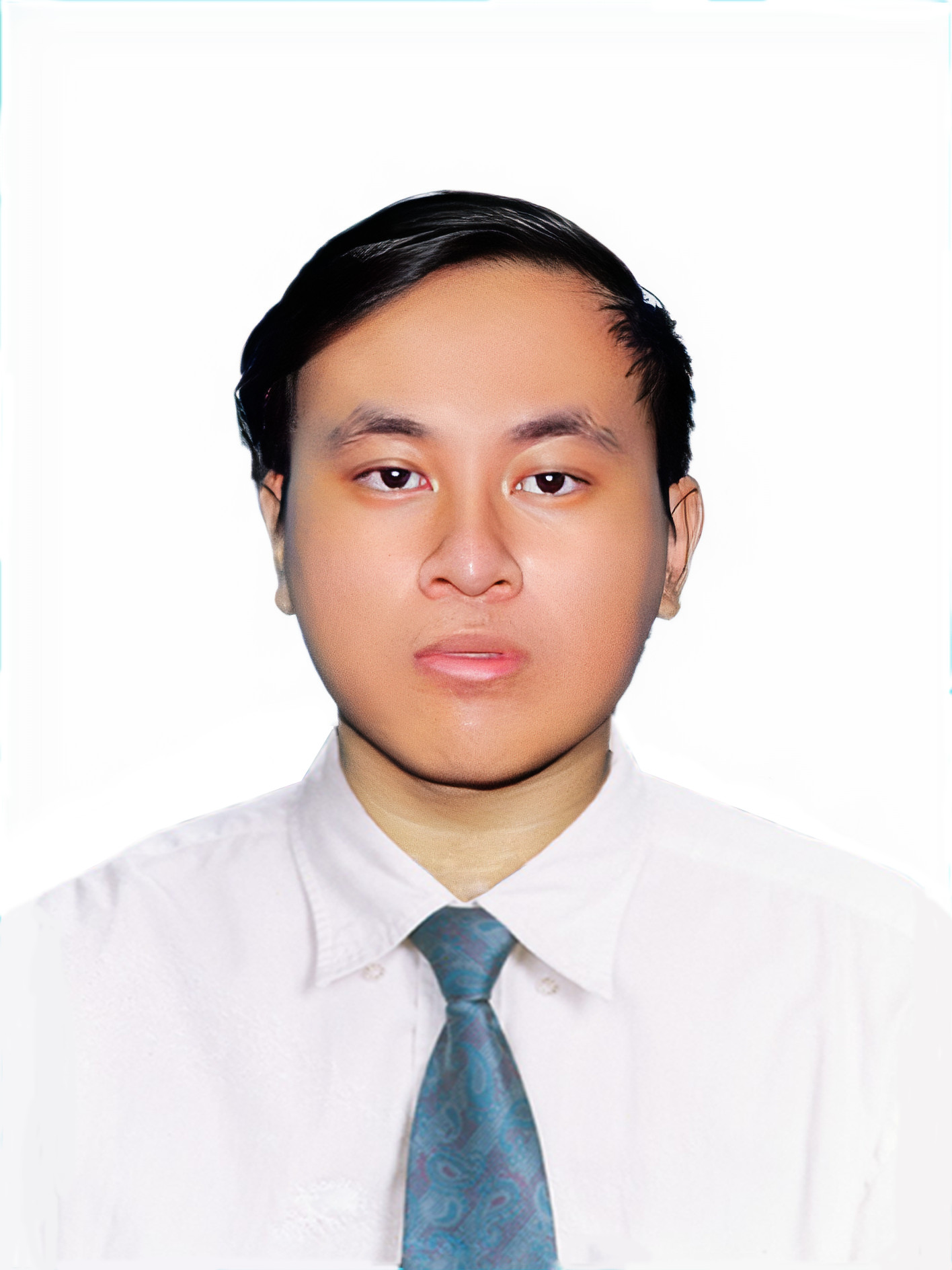}}]{Tuan Hai Vu} (Member, IEEE) received the B.S. degree in Software Engineering and the M.S. degree in Computer Science from the University of Information Technology, Vietnam National University, in 2021 and 2023, respectively. He is currently a Ph.D. student at the Architecture Lab at Nara Institute of Science and Technology, Japan. His research interests include quantum simulation acceleration and quantum machine learning.
\end{IEEEbiography}
\vskip -3.2\baselineskip plus -1fil
\begin{IEEEbiography}[{\includegraphics[width=1in,height=1.25in,clip,keepaspectratio]{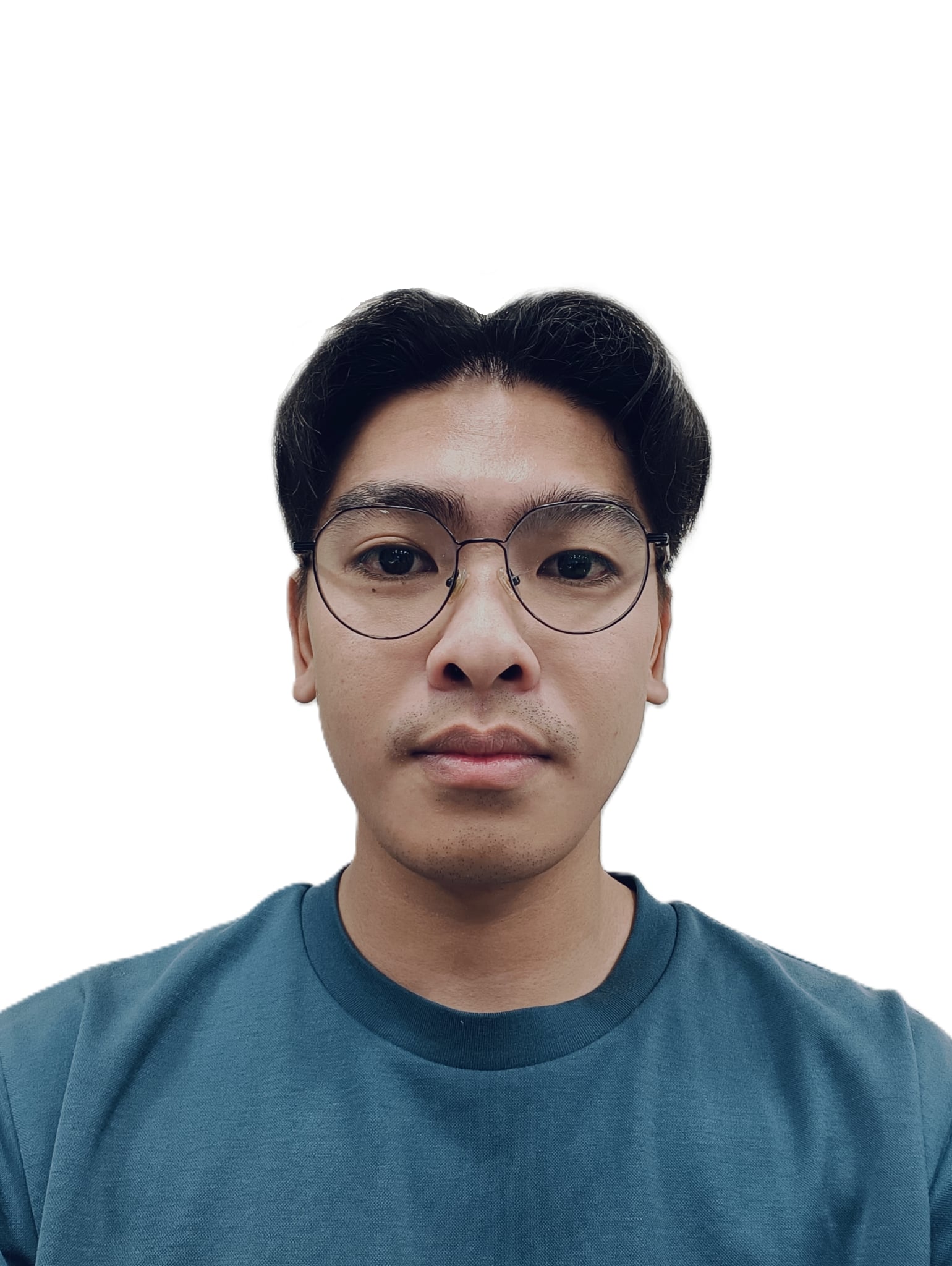}}]{Van Duy Tran} received the B.E. degree in IC and hardware design from Vietnam National University Ho Chi Minh City—University of Information Technology in 2022. He is currently pursuing the master’s degree in information science with Nara Institute of Science and Technology, Japan. His research interests include wireless communication systems, artificial intelligence, cryptography, ASICs, and VLSI design.
\end{IEEEbiography}
\vskip -3.2\baselineskip plus -1fil
\begin{IEEEbiography}[{\includegraphics[width=1in,height=1.25in,clip,keepaspectratio]{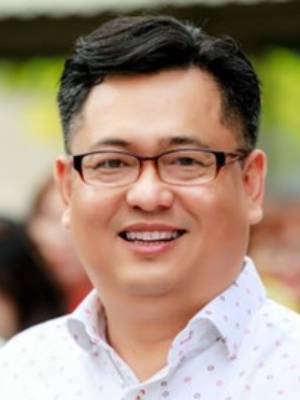}}]{Khang Nguyen} received his B.S. degree and M.S. degrees in Computer Science from University of Science, VNUHCM, Vietnam in 1990 and 1995. He received his Ph.D. degree in 2012 from the University of Science, VNU-HCM, Vietnam. He is currently the Vice-President of the University of Information Technology, VNU-HCM. His research interests include artificial intelligence and computer vision.
\end{IEEEbiography}
\vskip -3.3\baselineskip plus -1fil
\begin{IEEEbiography}[{\includegraphics[width=1in,height=1.25in,clip,keepaspectratio]{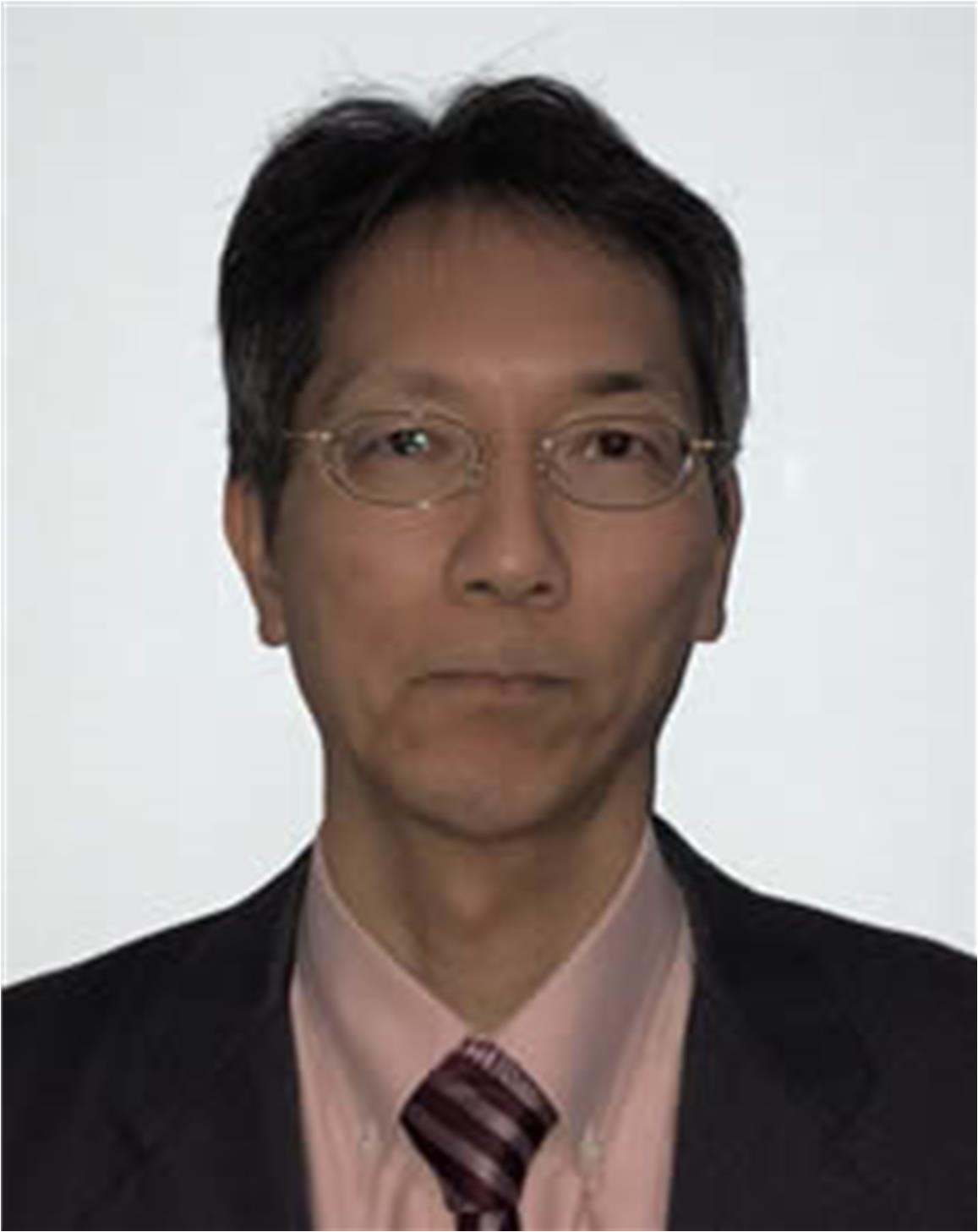}}]{YASUHIKO NAKASHIMA} (Senior Member, IEEE) received the B.E., M.E., and Ph.D. degrees in Computer Engineering from Kyoto University in 1986, 1988, and 1998, respectively. He was a Computer Architect with the Computer and System Architecture Department, FUJITSU Ltd., from 1988 to 1999. From 1999 to 2005, he was an Associate Professor with the Graduate School of Economics, Kyoto University. Since 2006, he has been a Professor with the Graduate School of Information Science, Nara Institute of Science and Technology (NAIST). His research interests include computer architecture, emulation, circuit design, and accelerators. He is a Fellow of IEICE.
\end{IEEEbiography}

\EOD

\end{document}